\documentclass{article}

\PassOptionsToPackage{numbers, sort&compress}{natbib}

\usepackage[preprint]{neurips_2025}

\usepackage[utf8]{inputenc} %
\usepackage[T1]{fontenc}    %
\usepackage{hyperref}       %
\usepackage{url}            %
\usepackage{booktabs}       %
\usepackage{amsfonts}       %
\usepackage{nicefrac}       %
\usepackage{microtype}      %
\usepackage{xcolor}         %

\usepackage{amsmath}
\usepackage{amssymb}
\usepackage{mathtools}
\usepackage{amsthm}

\usepackage{enumitem}
\usepackage{titlesec}

\usepackage{graphicx}
\usepackage{subcaption}
\usepackage{wrapfig}

\usepackage[capitalize,noabbrev]{cleveref}

\definecolor{LinkBlue}{HTML}{1A4FA3}
\hypersetup{
    colorlinks,
    linkcolor={LinkBlue},
    citecolor={LinkBlue},
    urlcolor={LinkBlue},
}
\usepackage[labelfont=bf]{caption}

\title{From Words to Amino Acids: Does the Curse of Depth Persist?}



\author{%
    \makebox[\textwidth][c]{%
        \textbf{Aleena Siji}\textsuperscript{\mdseries 1,2}\footnotemark[1] \hspace{.5pt} \footnotemark[3]\quad
        \textbf{Amir Mohammad Karimi Mamaghan}\textsuperscript{\mdseries 3}\footnotemark[1] \hspace{.5pt} \footnotemark[3]\quad
        \textbf{Ferdinand Kapl}\textsuperscript{\mdseries 1,2}\quad
        \textbf{Tobias Höppe}\textsuperscript{\mdseries 1,2}\quad
    }\\[0.25em]
    \makebox[\textwidth][c]{%
        \textbf{Emmanouil Angelis}\textsuperscript{\mdseries 1,2}\quad
        \textbf{Andrea Dittadi}\textsuperscript{\mdseries 1,2}\quad
        \textbf{Maurice Brenner}\textsuperscript{\mdseries 1,4}\quad
        \textbf{Michael Heinzinger}\textsuperscript{\mdseries 1,4}
    }\\[0.25em]
    \makebox[\textwidth][c]{%
        \textbf{Karl Henrik Johansson}\textsuperscript{\mdseries 3}\quad
        \textbf{Kaitlin Maile}\textsuperscript{\mdseries 5}\quad
        \textbf{Johannes von Oswald}\textsuperscript{\mdseries 5}\quad
        \textbf{Stefan Bauer}\textsuperscript{\mdseries 1,2}
    }\\[0.25em]
    \makebox[\textwidth][c]{%
        \textsuperscript{1}\,Technical University of Munich \quad
        \textsuperscript{2}\,Helmholtz AI, Munich \quad
        \textsuperscript{3}\,KTH Royal Institute of Technology \quad
    }\\[0.25em]
    \makebox[\textwidth][c]{%
        \textsuperscript{4}\,Institute of Computational Biology, Helmholtz Munich\quad
        \textsuperscript{5}\,Google, Paradigms of Intelligence Team
    }\\[0.25em]
}

\begin{document}

\maketitle
\footnotetext[1]{Equal contribution.}
\footnotetext[3]{Correspondence: \texttt{aleena.siji@helmholtz-munich.de,  amkm@kth.se}.}

\begin{abstract}
Protein language models (PLMs) have become widely adopted as general-purpose models, demonstrating strong performance in protein engineering and de novo design. Like large language models (LLMs), they are typically trained as deep transformers with next-token or masked-token prediction objectives on massive sequence corpora and are scaled by increasing model depth. 
Recent work on autoregressive LLMs has identified the \textit{Curse of Depth}: many later layers contribute little to the final output predictions. 
These findings naturally raise the question of whether a similar depth inefficiency also appears in PLMs, where many widely used models are not autoregressive, and some are multimodal, accepting both protein sequence and structure as input. 
In this work, we present a depth analysis of seven popular PLM families across model scales, spanning autoregressive, masked, and diffusion objectives, and quantify how layer contributions evolve with depth using a unified set of probing-, perturbation-, and downstream-evaluation measurements. 
Across models, we observe consistent depth-dependent patterns that extend prior findings on LLMs: a large fraction of task-relevant computation is concentrated in a subset of layers, while the remaining layers mainly provide incremental refinement of the final prediction. These trends persist beyond sequence-only settings and also appear in multimodal PLMs. Taken together, our results suggest that depth inefficiency is a common feature of modern PLMs, motivating future work on more depth-efficient architectures and training methods.\looseness=-1
\end{abstract}

\section{Introduction}
\label{sec:introduction}

Proteins are essential macromolecules in living organisms. They are specified by amino acid sequences, which largely determine a protein's three-dimensional structure, and in turn, its function. Recent progress in protein modeling has been driven by protein language models (PLMs), which are large-scale models trained on evolutionary-scale sequence databases using self-supervised objectives, and can capture meaningful biochemical and evolutionary properties even without explicit structural or functional information \citep{lin2023esm2, nijkamp2023progen2, ferruz2022protgpt2, wang2024dplm}. As a result, PLM representations can be used as general-purpose features for many downstream problems, including protein function annotation \citep{brandes2022proteinbert, rives2021esm1, elnaggar2021prottrans}, structure-related prediction tasks such as secondary structure and contact prediction \citep{rao2019evaluating, lin2023esm2}, and mutation effect prediction \citep{meier2021language, nijkamp2023progen2}. More recently, PLMs have rapidly moved beyond sequence-only encoders. Several recent models extend PLMs by incorporating additional modalities, most commonly the structural information, and are trained with objectives that enable multimodal generation \citep{hayes2025esm3, wang2025dplm2, geffner2025laproteina, chen2025apm}. This makes them useful not only as feature extractors but also as generative models which can be utilized for protein engineering and design.

This rapid progress has largely followed the same scaling and training paradigm that has shaped modern large language models (LLMs). In particular, PLMs and LLMs are typically built on the Transformer architecture \citep{vaswani2017attention} and trained with self-supervised language modeling objectives on massive sequence corpora: natural language text for LLMs and large protein sequence (and sometimes, structure) databases for PLMs \citep{rives2021esm1, lin2023esm2}. In both cases, the model learns contextual token representations by predicting tokens from context, either by masking random input tokens or by predicting the next token with causal masking. As a result, many of the practical questions that arise when scaling LLMs also naturally apply to PLMs.

As an example of such scaling questions, recent analyses of modern autoregressive LLMs suggest that simply stacking more transformer blocks does not necessarily translate into proportional gains in model capability. \citet{sun2025curse} formalize this as the \textit{Curse of Depth}: across several popular autoregressive LLM families, deeper layers often contribute much less than earlier layers, and pruning or perturbing many late layers causes only a small performance change. \citet{csordas2025language} reach a consistent conclusion by directly analyzing the residual stream. They find a sharp drop in layer contributions around the middle of the network, and show that skipping layers in the second half has a much smaller effect on subsequent computations, suggesting that late layers mainly refine the final output distribution rather than building reusable intermediate results. Together, these findings raise a broad concern for deep transformer models. In particular, increased depth may be used inefficiently, with a substantial fraction of layers being less effective during training and inference \citep{sun2025curse, csordas2025language}.

These findings in LLMs naturally raise the question of whether a similar depth inefficiency also appears in PLMs, where many widely used models are not autoregressive, and some are multimodal, incorporating additional information such as protein structure. While PLMs closely mirror LLMs in both architecture and training, this issue has not been systematically analyzed for PLMs. Furthermore, because of the key differences in modality and common training objectives, existing results for LLMs do not necessarily transfer to PLMs. In this work, we present a comprehensive depth analysis of a diverse set of widely used PLMs across model families, sizes, and training objectives, and quantify how layer contributions and representation changes evolve with depth using a unified suite of probing- and perturbation-based measurements. In particular, our contributions are as follows:

\begin{enumerate}[topsep=0pt]
    \item We present the first systematic study of depth usage in protein language models. We analyze 7 widely used PLM families with 26 model variants in total, covering autoregressive, masked, and diffusion objectives, and including two multimodal models.
    \item Building on recent depth analyses in LLMs \citep{csordas2025language}, we develop a unified depth-analysis framework for PLMs that combines probing- and perturbation-based measurements with layer-wise downstream evaluation on the popular ProteinGym benchmark \citep{notin2023proteingym}. Using this framework, we find a depth-usage pattern consistent with observations in LLMs: across all models, scaling leads to a depth inefficiency, where most of the downstream-relevant computation is formed in a subset of layers, often intermediate to later layers, and the remaining layers primarily provide incremental refinement of the final predictions.
    \item We show that this behavior generalizes beyond next-token prediction and also emerges under masked and diffusion objectives and persists in multimodal PLMs. By extending the analysis to structure-only and joint sequence-structure settings, we further show that similar patterns hold across modality streams, while cross-modal similarity analysis helps explain the distinct depth profile of ESM3. Together, these results suggest that depth inefficiency is a general property of modern PLMs and motivate more depth-efficient architectures and training methods.
\end{enumerate}

\section{Related Work}
\label{sec:related_work}

\paragraph{Protein Language Models.}
In recent years, alongside the rapid progress of LLMs, PLMs have also evolved quickly in terms of architecture, modality, and training objectives. Early large-scale sequence models showed that transformers trained on billions of amino acids learn transferable representations of protein grammar for diverse downstream tasks~\citep{brandes2022proteinbert, elnaggar2021prottrans}. Building on this, the ESM family scaled bidirectional transformer encoders from ESM-1b~\citep{rives2021esm1} to ESM2~\citep{lin2023esm2}, improving single-sequence representations and enabling strong structure-related predictions. More recently, PLMs have expanded beyond sequence-only inputs: \citet{hayes2025esm3} propose ESM3, trained over sequence, structure, and function, enabling conditional generation over any modality. In parallel, PLMs have diversified in training objectives. Autoregressive models such as the ProGen family~\citep{madani2020progen, nijkamp2023progen2, bhatnagar2025progen3} and ProtGPT2~\citep{ferruz2022protgpt2} support controllable sequence generation. Most recently, \citet{bhatnagar2025progen3} introduce ProGen3, a Mixture-of-Experts (MoE) autoregressive PLM family that scales up model size and training, and show that this scaling improves the quality and diversity of generated proteins. Meanwhile, \citet{wang2024dplm} introduce discrete diffusion for protein sequences and \citet{wang2025dplm2} extend it with structure for joint sequence-structure modeling in protein design. Finally, retrieval-augmented approaches such as Profluent-E1~\citep{jain2025e1} leverage homolog retrieval to improve zero-shot predictions. Together, these lines of work reflect a broader shift toward multimodal, generative, and retrieval-enhanced PLMs for both prediction and design. 

\paragraph{Depth-wise Analysis of Transformers.}
Depth analysis of Transformer-based models has evolved from probing what layers encode to intervening on layers to test their contribution to the output. Early works used probes and attention analyses to map linguistic structure across depth, revealing layer-wise differences in what is linearly recoverable from representations \citep{hewitt2019structural, tenney2019you, clark2019does}. More recent intervention studies show that autoregressive language models can tolerate dropping or swapping many deeper layers, motivating stage-like views of inference \citep{lad2024remarkable}. Complementing this, \citet{skean2025layer} identify mid-layer bottlenecks and show that intermediate layers can outperform the final layer on many embedding-style tasks. Two recent works then consolidate these observations into a depth-inefficiency view: \citet{csordas2025language} find a mid-depth transition after which later layers have much smaller downstream influence, and \citet{sun2025curse} formalize this as the \textit{Curse of Depth} and propose Layer Norm Scaling to improve deeper-layer learning.
Inspired by these findings, in contrast to prior works that are mostly focused on autoregressive LLMs, we ask whether a similar depth inefficiency arises in PLMs. We study layer contributions across autoregressive, masked, and diffusion PLMs, and include multimodal models.

\section{Problem Setup}
\label{sec:problem_setup}

In this section, we briefly describe the PLM families included in our study and the evaluation setting used to assess depth efficiency across models and scales.

\paragraph{Models.}
Our model set is chosen to reflect the main training paradigms and to cover both sequence-only and multimodal settings. In particular, the selected models vary along two main axes that are central to our study: i) %
\textbf{training objectives}, namely masked-token prediction, autoregressive next-token prediction, and discrete diffusion, and ii) %
\textbf{input modality}, ranging from sequence-only encoders to models that incorporate additional protein signals such as structural and functional information.
As sequence-only PLMs, we include ESM2 \citep{lin2023esm2}, a widely used family of masked encoders that serves as a standard baseline, DPLM \citep{wang2024dplm}, Profluent-E1 \citep{jain2025e1}, ProGen2 \citep{nijkamp2023progen2}, and ProGen3 \citep{bhatnagar2025progen3}. Notably, while E1 is sequence-only, it is trained in a retrieval-augmented setting by conditioning on additional unaligned homolog sequences. As multimodal PLMs, we include ESM3 \citep{hayes2025esm3}, a masked model trained on protein sequence, structure, and function, and DPLM2 \citep{wang2025dplm2}, an upgraded version of DPLM trained jointly on protein sequence and structure. Furthermore, to enable consistent comparisons across all families, we first conduct our analysis on the sequence input pathway for every model, including multimodal models, and separately analyze the structure and multimodal pathways of the multimodal models afterwards. Finally, for each model family, we consider all publicly available sizes, allowing us to quantify how depth-dependent effects evolve with scale. A full description of the models is provided in \cref{app:models}.

It is worth noting that some model families also differ in their pretraining datasets and data scales, which could, in principle, influence the observed behavior. Since our goal is not to benchmark or rank models against each other, we do not attempt to control for or attribute effects to these data differences. Instead, we primarily focus on depth-dependent patterns \textit{within} each model family by comparing different sizes within the same family, and treat cross-family comparisons as out of scope. Importantly, within a given model family, models are typically trained on the same pretraining dataset with broadly comparable training setups and often similar token budgets, which helps isolate depth-related trends.

\paragraph{Evaluations.}
Our primary analysis follows the probing- and intervention-based depth analysis of \citet{csordas2025language}. We measure layer contributions via controlled layer-skipping perturbations and their downstream effects on subsequent computations, and complement this with layer--layer similarity analysis to track how representations evolve across depth. We further study how the model’s implied output distribution changes across layers using layer-wise output readouts. We apply these analyses not only to the sequence stream, but also to the structure-only and joint sequence-structure (multimodal) inputs for the multimodal PLMs, enabling a modality-conditional view of depth efficiency within these models. 
Finally, to connect these intrinsic measurements to practical utility, we evaluate layer-wise downstream performance on ProteinGym \citep{notin2023proteingym}, a standard benchmark for zero-shot mutation-effect prediction based on deep mutational scanning assays. Further details are provided in \cref{sec:experiments} and \cref{app:evaluation_setup}.

\paragraph{Limitations.}
While our goal is to provide a robust and informative study of depth efficiency in PLMs, our analysis has limitations in model coverage and evaluation scope. First, our multimodal coverage is limited to two model families (ESM3 and DPLM2). Although we extend our probing and intervention-based analysis to sequence-only, structure-only, and joint sequence-structure inputs for these models, broader coverage across additional multimodal and structure-aware PLMs would strengthen the generality of conclusions about modality-specific effects. Second, our downstream evaluation focuses on ProteinGym, which provides a biologically meaningful and standardized benchmark for zero-shot mutation-effect prediction, but does not cover the full range of downstream settings in which PLMs are used. Finally, our evaluation is based on randomly sampled proteins without explicitly controlling for sequence length, and depth-efficiency patterns may therefore differ for substantially longer sequences or for protein subsets with different length distributions than those represented in our evaluation.

\section{Experiments}
\label{sec:experiments}

We conduct a set of probing- and intervention-based experiments to analyze depth usage from different perspectives. First, we intervene on each layer and measure its contribution to downstream computations, complemented by layer-layer similarity analysis of representations (\cref{subsec:skiplayer_effects}). Next, we study how the model’s prediction distribution evolves with depth by extracting a layer-wise implied token distribution and comparing it to the final-layer distribution (\cref{subsec:lens}). We then evaluate downstream utility via layer-wise performance on ProteinGym and analyze when strong predictive signals emerge across the network (\cref{subsec:proteingym}). 
Finally, we extend our analyses to multimodal PLMs by studying structure-only and joint sequence-structure inputs, and use cross-modal similarity analysis to better understand the distinct depth-wise behavior of ESM3 (\cref{subsec:multimodal}). Unless stated otherwise, experiments in \cref{subsec:skiplayer_effects,subsec:lens} use 40 protein sequences randomly sampled from UniRef50 \citep{suzek2007uniref} as input prompts, while the multimodal analyses in \cref{subsec:multimodal} use 50 proteins randomly sampled from the Protein Data Bank (PDB). Full experimental details are provided in \cref{app:evaluation_setup}. We focus on ESM2 as a representative and widely used PLM in the main text, and report results for the other models in \cref{app:additional_results}.

\subsection{How much does each layer contribute to later computations?} 
\label{subsec:skiplayer_effects}
We measure how much each layer contributes to \emph{later} computations by skipping that layer for earlier token positions and tracking how the intervention propagates downstream. Concretely, we sample a split position and run a second forward pass, skipping the chosen source layer only up to the split. We then quantify the resulting change in subsequent-layer representations on the remaining (future) token positions and report the propagated effects. In addition, we analyze how representations evolve across depth by computing layer-layer representation similarity. Concretely, for each pair of layers, we compute the similarity between their token representations using linear CKA, a standard metric for representation similarity \citep{kornblith2019cka}. We report these layer-layer similarity matrices alongside the skiplayer heatmaps to provide a complementary, non-interventional view of depth-wise representation change. See \cref{app:skiplayer_details} for more details.

For the skiplayer analysis, the original setup of \citet{csordas2025language} is tailored to autoregressive next-token prediction. For ProGen2 and ProGen3, we therefore follow the same procedure, skipping a layer only up to a randomly sampled token position and measuring its effect on future tokens. For masked and diffusion-based models that are not trained with a causal left-to-right objective, we adapt the notion of “future tokens” to capture cross-positional effects as follows: we mask 15\% of tokens, intervene on a random subset of token positions, and measure effects on the remaining non-intervened masked tokens. Specifically, we sample 20--80\% of masked positions and 20--80\% of non-masked positions for intervention, and compute effects on the remaining (i.e., non-intervened) masked positions. We also measured effects on all non-intervened tokens (i.e., masked and non-masked), and observed similar trends. Hence, we report results only for the former.

\begin{figure*}[t]
    \centering
    \includegraphics[width=0.98\textwidth]{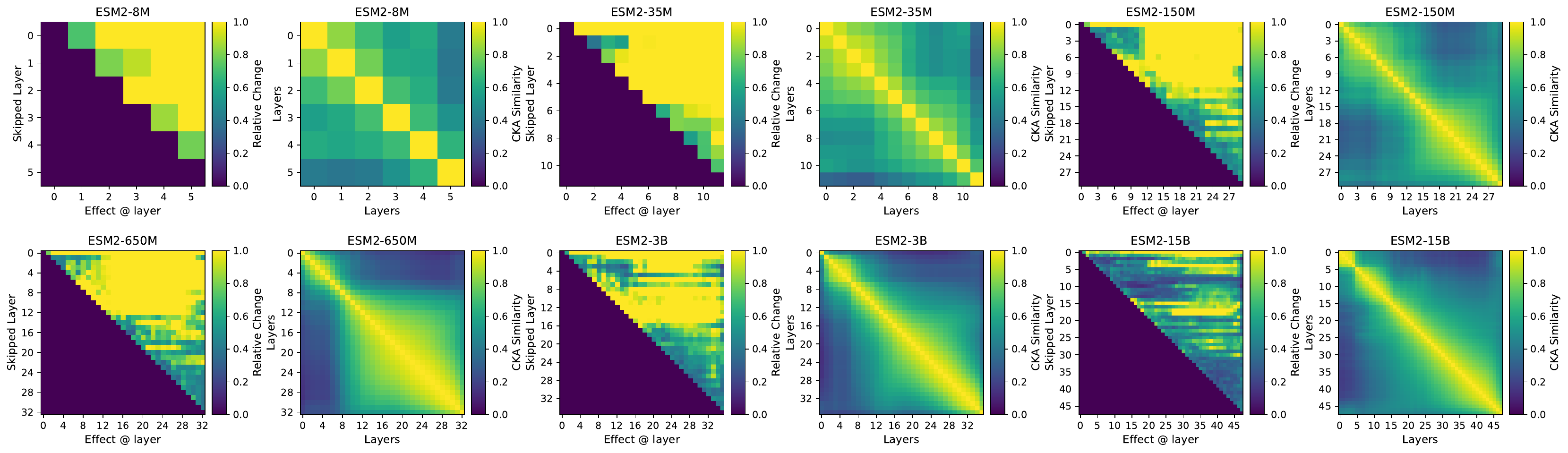}
    \caption{\textbf{Maximum propagated effect of skipping each layer on future-token computations, alongside layer-layer CKA similarity, in ESM2.} Here, “future” refers to a held-out subset of non-intervened masked tokens. Even at 35M, skipping later layers produces relatively weak propagated effects compared to skipping early layers. From 150M onward, this separation becomes clear: a substantial fraction of late layers can be skipped with only minor effects on subsequent computations on future tokens. Moreover, the low-effect regions also report high layer-layer similarity. This pattern strengthens with scale, indicating that downstream sensitivity increasingly concentrates in early-to-middle layers, while later layers mainly refine the final prediction. We also observe localized low-effect regions among early layers. This aligns with the stage-wise view of \citet{lad2024remarkable} and suggests that depth is organized into multiple inference stages with weaker dependencies across certain layer ranges.
    }
    \label{fig:skiplayer_layers_ESM2}
\end{figure*}

\paragraph{Results.}
\cref{fig:skiplayer_layers_ESM2} and \cref{fig:skiplayer_layers_E1,fig:skiplayer_layers_DPLM,fig:skiplayer_layers_DPLM2,fig:skiplayer_layers_ESM3,fig:skiplayer_layers_ProGen2,fig:skiplayer_layers_ProGen3} in \cref{app:skiplayer_effects} show the results for each model family. Across all models, we observe a consistent pattern: skipping earlier layers typically causes larger changes in later layers, whereas skipping later layers often has only a minor effect on the remaining computation. The low-effect region also reports high layer-layer similarity, providing additional evidence that later layers only minimally refine the final predictions. For ESM2 (\cref{fig:skiplayer_layers_ESM2}), this trend is visible already at 35M, becomes clear from 150M onward, and strengthens with scale. At 15B, skipping a large fraction of layers in the second half of the model has little effect on subsequent layers. 

DPLM (\cref{fig:skiplayer_layers_DPLM}) shows a similar pattern starting from the smallest size (150M), and it becomes clearer with scale. DPLM2 (\cref{fig:skiplayer_layers_DPLM2}) behaves similarly across all sizes. Profluent-E1 (\cref{fig:skiplayer_layers_E1}) follows the same overall trend, but with a somewhat different structure in how depth regions most strongly influence later layers. For ESM3 (\cref{fig:skiplayer_layers_ESM3}), the trend is somewhat different from the other models: skipping layers in the first half of the network tends to produce relatively small propagated changes, while a low-effect regime in late layers is still present but less sharply separated. We investigate ESM3’s depth-wise trends further in \cref{subsec:esm3_modality_alignment}. Finally, among the autoregressive models, ProGen2 (\cref{fig:skiplayer_layers_ProGen2}) reports the same overall pattern already in the small variant and it becomes more pronounced with scale. ProGen3 (\cref{fig:skiplayer_layers_ProGen3}), which is designed as a sparse MoE model, also shows the same overall pattern, but it does not become more pronounced with scale. This is consistent with recent findings in LLMs suggesting that sparsity can improve depth utilization \citep{muhtar2026does}. In addition to the layer-wise effects, we also quantify how skipping each layer changes the final model outputs by measuring the change in output norm before and after the intervention, again restricted to future tokens. We report and discuss the results in \cref{app:additional_results}.

Interestingly, for all non-autoregressive models (i.e., ESM2, ESM3, DPLM, DPLM2, and Profluent-E1), we observe two low-effect regions: one in the early layers and one in the later layers. This is evident both in the skiplayer effects and in the layer-layer similarity heatmaps, and suggests that depth may be organized into stages with weaker dependencies across certain layer ranges. This is consistent with the findings of \citet{lad2024remarkable}, which show that for autoregressive LLMs, transformer computation can be decomposed into several stages of inference. The skiplayer effect heatmaps suggest that similar stage-like patterns may also arise in masked and diffusion PLMs. We hypothesize that the early layers may be learning and refining different features with low cross-positional effects at that stage, but having high effects in later computation stages. We leave further investigation to future work.

\begin{figure*}[t]
    \centering
    \includegraphics[width=0.9\textwidth]{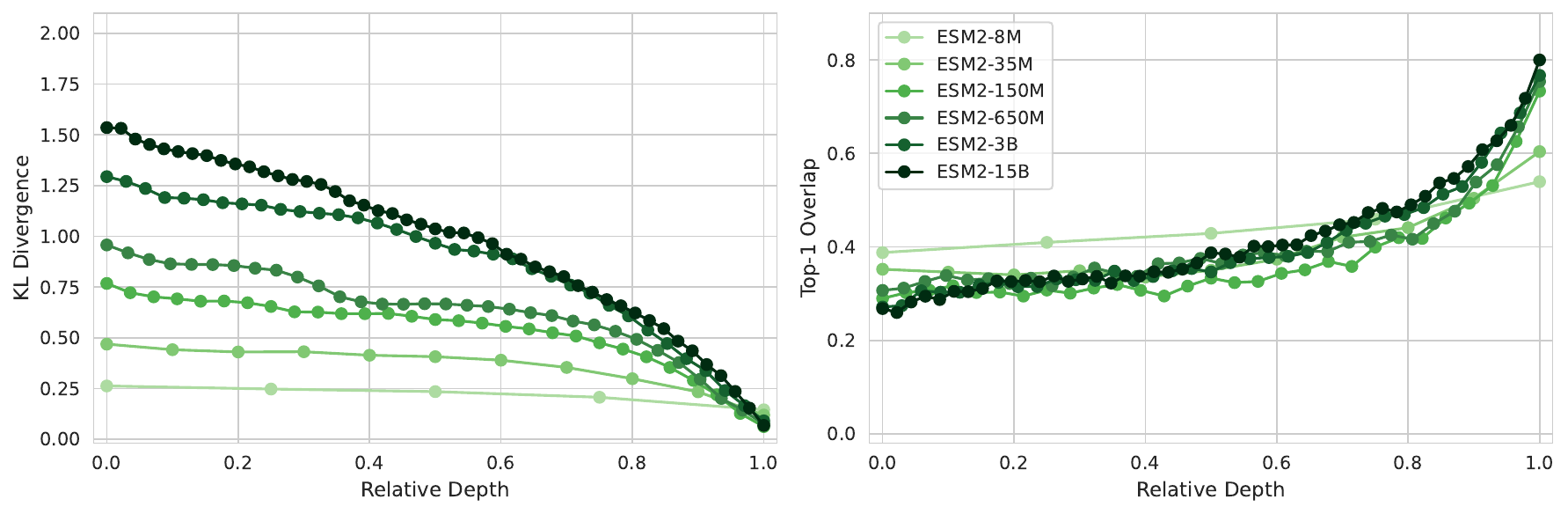}
    \caption{\textbf{TunedLens analysis for ESM2 across depth: KL divergence to the final output distribution (left) and top-1 overlap with the full-model prediction (right).} Across ESM2 variants, KL divergence slowly decreases toward later layers, indicating that deeper layers make increasingly incremental updates that bring the distribution closer to the final output. Consistently, top-1 overlap is lower in early layers and increases toward the end of the network, with the most pronounced gains in the final layers for larger variants. Together, these trends suggest that the top prediction becomes increasingly stable with depth, while later layers mainly refine confidence over tokens rather than change the predicted token.}
    \label{fig:tunedlens_ESM2}
\end{figure*}

\paragraph{Summary.}
Across all models and training objectives, we observe a consistent trend: as models scale, the propagated effects of layer skipping are dominated by early-to-middle layers, whereas skipping many late layers only minimally changes subsequent computations. This trend is further supported by high layer-layer similarity among later layers. ESM3 is a notable exception, where the weakest effects (and high-similarity regions) are mainly concentrated in the earliest layers rather than later, although it still shows a weaker late-layer regime. ProGen3 also provides an interesting nuance: while it follows the same overall trend, the separation does not become more pronounced with scale, which may be related to its sparse MoE design and is consistent with recent work suggesting that sparsity can mitigate the curse of depth in LLMs \citep{muhtar2026does}. Overall, these findings mirror those in LLMs and suggest a form of depth inefficiency, in which layers with weak propagated effects act mainly as incremental refinement steps on the final output rather than major drivers of subsequent computation \citep{lad2024remarkable, sun2025curse, csordas2025language, gromov2024unreasonable}.\looseness=-1

\subsection{How does the output distribution evolve across depth?}
\label{subsec:lens}

We further study how a model’s output distribution evolves with depth using layer-wise output readouts based on LogitLens and TunedLens \citep{logitlens,belrose2023eliciting}. Concretely, after each layer, we take the residual stream, apply normalization and unembedding to obtain token logits, and interpret the resulting softmax as the layer’s implied prediction distribution. We consider two variants that differ only in the prediction head: in LogitLens, we use the model’s original final normalization and unembedding at every layer, whereas in TunedLens \citep{belrose2023eliciting}, we use a separate prediction head for each layer that is tuned on a subset of UniRef50, which improves the alignment between intermediate representations and the output space. Unless stated otherwise, we use TunedLens for the sequence-stream results in this section, and use LogitLens for the multimodal analyses in \cref{subsec:multimodal}. We then report the KL divergence between the layer-wise distribution and the final distribution produced by the full model. In addition, we measure how often the top-$k$ tokens under the layer-wise distribution overlap with the top-$k$ tokens under the final distribution. Since protein vocabularies are much smaller than natural-language vocabularies and are essentially limited to the standard amino-acid alphabet ($|\mathcal{V}| \approx 20$), we report top-$k$ overlap with $k=1$. Furthermore, to adapt this to masked and diffusion-based models, we use the same setup as in \cref{subsec:skiplayer_effects} and compute results for masked positions. For further details, see \cref{app:lens_details}.

\paragraph{Results.} 
The TunedLens results are shown in \cref{fig:tunedlens_ESM2} and \cref{fig:tunedlens_ESM3,fig:tunedlens_E1,fig:tunedlens_DPLM,fig:tunedlens_DPLM2,fig:tunedlens_ProGen2,fig:tunedlens_ProGen3} in \cref{app:lens}. Overall, the KL-divergence curves show a consistent trend across all model families: deeper layers produce layer-wise distributions that progressively approach the model’s final distribution. For ESM2 and ESM3, this behavior is visible already at smaller scales and becomes clear from 150M onward in ESM2, where the KL divergence steadily decreases toward later layers, and a similar decreasing trend is also observed for ESM3 (\cref{fig:tunedlens_ESM2,fig:tunedlens_ESM3}). Profluent-E1 likewise exhibits a clear downward trend across all sizes (\cref{fig:tunedlens_E1}). For the diffusion-based models, both DPLM and DPLM2 show the same overall pattern, with KL divergence decreasing toward later layers across sizes, indicating increasingly incremental distribution-level refinement in deeper layers (\cref{fig:tunedlens_DPLM,fig:tunedlens_DPLM2}). Finally, for the ProGen family, ProGen2 shows a consistent profile; from ProGen2-medium onward, the KL divergence is relatively steady in early depth and then gradually decreases toward later layers, and ProGen3 follows the same overall trend (\cref{fig:tunedlens_ProGen2,fig:tunedlens_ProGen3}).

The top-1 overlap curves provide a complementary view of the same refinement behavior. Across model families, agreement with the final prediction is often already relatively high in early layers and remains fairly stable through much of the network, while the KL divergence stays stable or decreases only slightly at first and then declines more clearly toward later layers. Taken together, these results suggest that the top prediction is often fixed well before the last layer, while later layers mainly refine the confidence over tokens and align the full distribution with the final output.


\paragraph{Summary.}
The results are consistent with prior depth analyses in LLMs \citep{sun2025curse, csordas2025language, skean2025layer, gromov2024unreasonable} and align with our observations in \cref{subsec:skiplayer_effects}. Across model families, later layers tend to make smaller, incremental updates that mainly refine an increasingly stable output distribution, rather than inducing large shifts in the model’s predictions, and this refinement behavior becomes clearer with scale in many cases.

\subsection{How does downstream performance vary across depth?}
\label{subsec:proteingym}

To further analyze the contributions of different layers, we connect the intrinsic measurements in \cref{subsec:skiplayer_effects,subsec:lens} to practical utility by evaluating layer-wise performance on ProteinGym \citep{notin2023proteingym}. ProteinGym is a large-scale benchmark for mutation-effect prediction that includes 217 standardized deep mutational scanning (DMS) assays spanning millions of mutated sequences across diverse proteins and phenotypes. Following the official ProteinGym evaluation framework, we score variants in a zero-shot manner, compute Spearman’s rank correlation between predicted scores and experimental measurements for each assay, and report the average Spearman correlation across all assays.

\begin{wrapfigure}[22]{r}{0.5\textwidth}
    \vspace{-1.5em}
    \centering
    \includegraphics[width=\linewidth]{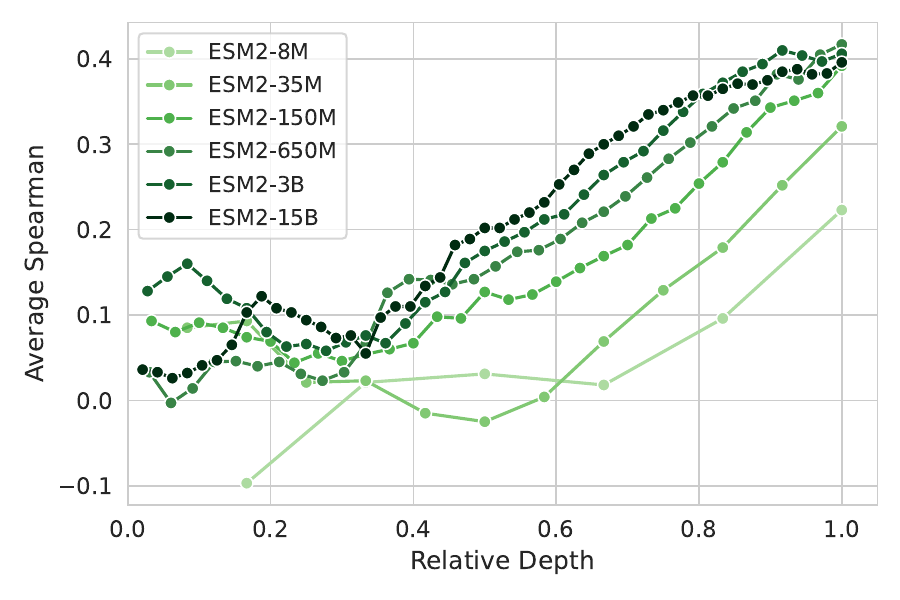}
    \caption{\textbf{Layer-wise ProteinGym performance for ESM2.} Average Spearman correlation as a function of relative depth, normalized to $[0,1]$, where predictions are taken from each layer. Performance improves with depth for all model sizes, but the largest models exhibit diminishing returns in the final layers, suggesting that earlier layers already capture much of the signal and later layers mainly provide small refinements to the final predictions.}
    \label{fig:layerwise_average_spearman_ESM2}
\end{wrapfigure}

For our experiments, we utilize the official ProteinGym codebase. For each layer, we compute mutation scores from that layer’s logits using the model’s standard prediction head, without additional fine-tuning. For ESM2, ESM3, and Profluent-E1, we use the masked-marginal scoring procedure \citep{meier2021language}. For diffusion-based models, i.e., DPLM and DPLM2, we apply the same masked-marginal protocol. For the autoregressive ProGen2 and ProGen3, we use the likelihood-ratio scoring provided by the official ProteinGym implementation. For more details, see \cref{app:proteingym_details}.

\paragraph{Results.}
\cref{fig:layerwise_average_spearman_ESM2} and \cref{fig:layerwise_average_spearman_DPLM,fig:layerwise_average_spearman_DPLM2,fig:layerwise_average_spearman_E1,fig:layerwise_average_spearman_ESM3,fig:layerwise_average_spearman_ProGen2,fig:layerwise_average_spearman_ProGen3} (with full phenotype breakdowns in \cref{fig:layerwise_full_ESM2,fig:layerwise_full_ESM3,fig:layerwise_full_E1,fig:layerwise_full_DPLM,fig:layerwise_full_DPLM2,fig:layerwise_full_ProGen2,fig:layerwise_full_ProGen3}) in \cref{app:proteingym} summarize the downstream results. Across all models except for ESM3, performance generally improves with depth, but for sufficiently large models, most gains happen in early-to-middle layers, after which additional layers yield noticeably smaller improvements. The trend is also not always strictly monotonic: for some smaller variants and occasionally even in early depth for larger ones, intermediate layers can temporarily underperform shallower layers before performance recovers, suggesting a short adjustment phase before later-layer refinements.

In particular, for ESM2 (\cref{fig:layerwise_average_spearman_ESM2}), a clear ``knee'' emerges around the 650M scale: smaller variants continue to improve fairly steadily across depth, whereas larger models (650M and above) reach strong performance by middle layers and then show diminishing gains towards the final layers. Furthermore, across all sizes, the first half of the layers does not improve the downstream performance much, which further supports the existence of several stages of inference \citep{lad2024remarkable} in PLMs. DPLM and DPLM2 (\cref{fig:layerwise_average_spearman_DPLM,fig:layerwise_average_spearman_DPLM2}, respectively) exhibit a similar pattern with earlier onset: even at 150M, we observe a mild saturation which becomes clearer as the model size increases. For Profluent-E1 (\cref{fig:layerwise_average_spearman_E1}), the saturation regime is present across all sizes, with most gains achieved before the last 40\% of the layers. For ProGen2 (\cref{fig:layerwise_average_spearman_ProGen2}), the plateau becomes clear from ProGen2-medium onward and is especially pronounced for large and xlarge variants, where performance changes little after roughly the first half of the network. For ProGen3 (\cref{fig:layerwise_average_spearman_ProGen3}), performance improves more steadily across depth across sizes, with a less pronounced late-layer plateau compared to ProGen2. Finally, for ESM3 (\cref{fig:layerwise_average_spearman_ESM3}), the depth profile is different: performance is largely flat over roughly the first 60\% of layers, with most of the gains concentrated in the later layers. This is consistent with our earlier probing and intervention results in \cref{subsec:skiplayer_effects} for ESM3 (\cref{fig:skiplayer_layers_ESM3,fig:skiplayer_outnorm_ESM3}), which also suggested a distinct depth-wise pattern, with weaker effects in the earliest layers and stronger contributions emerging around mid-to-late layers. We further analyze ESM3's behavior in \cref{subsec:esm3_modality_alignment}.

\paragraph{Summary.} 

For sufficiently large PLMs, intermediate layers often capture most of the signal required for strong downstream performance, while later layers provide smaller, incremental gains. For ESM3, downstream gains emerge mainly in later layers, consistent with its distinct depth-wise patterns in our probing and intervention analyses. Overall, these results complement our intrinsic measurements and support our main conclusion: across diverse PLM families and training objectives, model scaling is associated with depth inefficiency, where downstream-relevant computation tends to concentrate in a subset of layers (often intermediate to later layers), and the remaining layers provide more incremental refinement of the final prediction.

\subsection{How does depth usage change across modalities?}
\label{subsec:multimodal}

\subsubsection{Do structure-only and multimodal inputs show the same depth inefficiency?}
\label{subsec:multimodal_streams}

So far, we have mainly analyzed the sequence pathway of each model. To test whether the observed depth-inefficiency trends extend to other modality settings, we repeat the same analyses from \cref{subsec:skiplayer_effects,subsec:lens} for the two multimodal PLMs in our study, ESM3 and DPLM2, under (i) a \emph{structure-only} input stream and (ii) a \emph{multimodal} setting where both sequence and structure are provided. Concretely, we run skiplayer intervention, layer-layer similarity analysis, and LogitLens readouts, but using structural tokens as inputs (structure-only) or jointly conditioning on both modalities (sequence-structure). For these experiments, we randomly sample 50 proteins from the Protein Data Bank (PDB). Full experimental details and additional results are provided in \cref{app:multimodal}.\looseness=-1

\begin{figure*}[t]
    \centering
    \begin{minipage}[t]{0.44\textwidth}
        \centering
        \includegraphics[width=\textwidth]{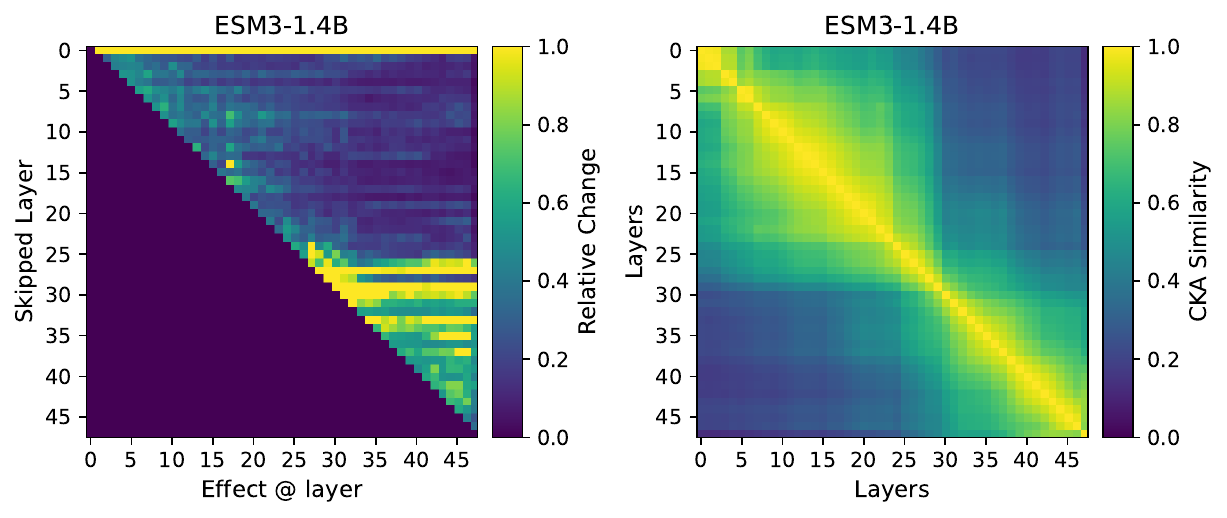}
    \end{minipage}\hfill
    \begin{minipage}[t]{0.54\textwidth}
        \centering
        \includegraphics[width=\textwidth]{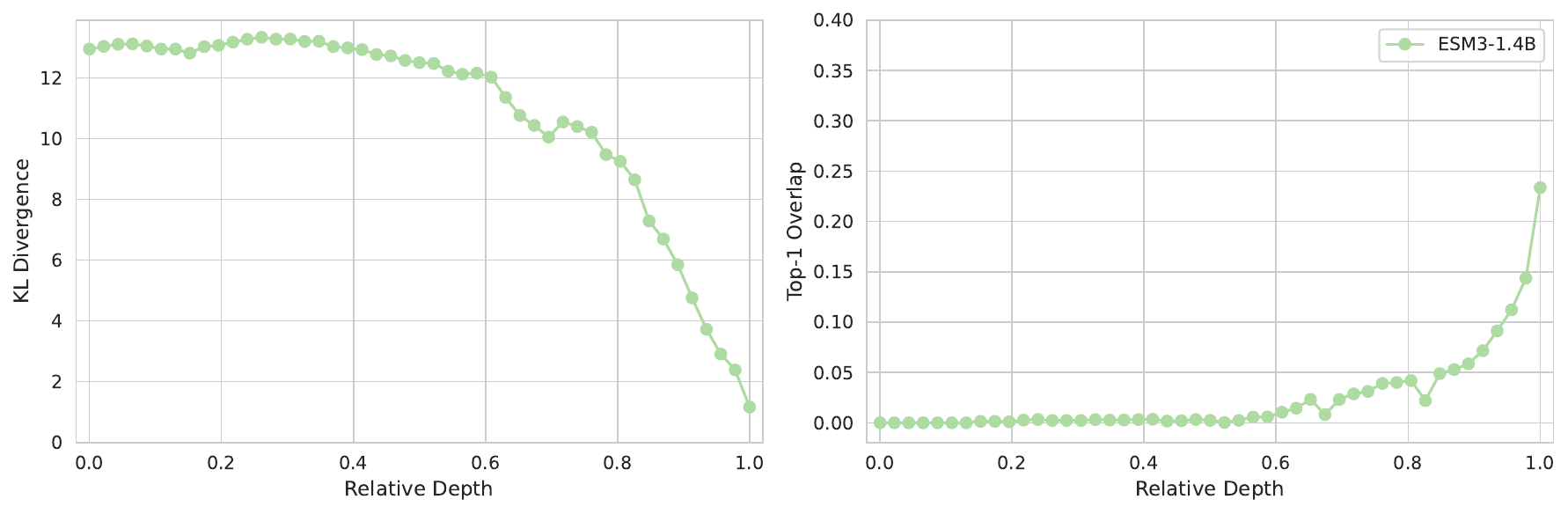}
    \end{minipage}

    \caption{\textbf{Depth analysis results for ESM3 for the structure stream.} \emph{Left:} skiplayer propagation and layer-layer similarity, and \emph{Right:} LogitLens readouts. Aligned with the sequence-stream results, ESM3 shows weaker propagated effects in early layers, with stronger contributions emerging in mid-to-late layers.}
    \label{fig:skiplayer_logitlens_structure_esm3}
\end{figure*}

\paragraph{Results.}
The structure- and multimodal-stream results are shown in \cref{fig:skiplayer_logitlens_structure_esm3} and in \cref{fig:skiplayer_layers_structure_DPLM2,fig:skiplayer_layers_multimodal_ESM3,fig:skiplayer_layers_multimodal_DPLM2,fig:logitlens_structure_dplm2,fig:logitlens_multimodal_dplm2,fig:logitlens_multimodal_esm3} in \cref{app:multimodal}. Across both multimodal PLMs and the structure-only and multimodal settings, we observe the same high-level trend as in the sequence pathway for each model. In several cases, these depth-inefficiency patterns are even more pronounced when conditioning on structure or on both modalities, with a clearer separation between high-effect and low-effect regimes in the skiplayer and similarity heatmaps. In particular, for ESM3, we again observe weaker late-layer effects overall in both settings, but the transition is less uniform, with a subset of mid-to-late layers still showing some influence (\cref{fig:skiplayer_logitlens_structure_esm3,fig:skiplayer_layers_multimodal_ESM3,fig:logitlens_multimodal_esm3}). 
For DPLM2, both the structure-only and sequence-structure settings show a sharp separation between early/mid layers with strong propagated effects and late layers with much weaker effects (\cref{fig:skiplayer_layers_structure_DPLM2,fig:skiplayer_layers_multimodal_DPLM2,fig:logitlens_structure_dplm2,fig:logitlens_multimodal_dplm2}). The LogitLens results are consistent with this picture, with KL divergence decreasing toward later layers and top-1 overlap increasing.

\subsubsection{Why does ESM3 exhibit a different depth profile?}
\label{subsec:esm3_modality_alignment}

To better understand why ESM3 exhibits a distinct depth-wise behavior compared to other PLMs, we analyze how its \emph{sequence-only} and \emph{structure-only} representations evolve across depth. Concretely, for the same set of proteins used in \cref{subsec:multimodal_streams}, we compute (i) the sequence-stream representations obtained when providing only sequence to the model, and (ii) the structure-stream representations obtained when providing only structure. We then compute a cross-modal layer-layer similarity matrix that measures how similar the sequence representation from a given layer is to the structure representation from another layer for the same protein. Here, we use cosine similarity rather than linear CKA, since our goal is to measure direct cross-modal alignment between per-protein representations, rather than similarity between representation spaces as a whole. This analysis provides a direct view of whether early depth is primarily devoted to aligning the two modalities into a shared representation space, or whether both streams develop independently across depth.

\begin{wrapfigure}[28]{r}{0.42\textwidth}
    \vspace{-1.5em}
    \centering
    \includegraphics[width=\linewidth]{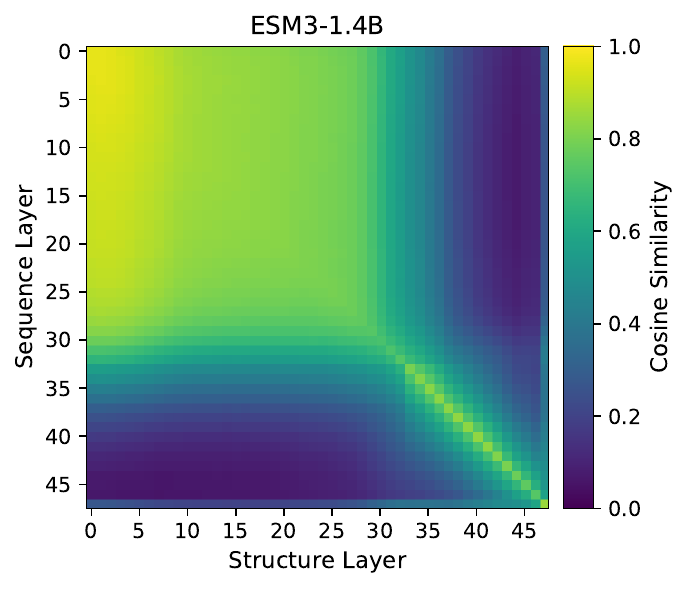}
    \caption{\textbf{Cross-modal layer--layer cosine similarity between sequence-only and structure-only representations for ESM3.} Each cell compares a sequence-layer representation to a structure-layer representation for the same proteins, with similarities averaged across proteins. The first $\sim$60\% of layers show high cross-modal similarity, suggesting strong early sequence-structure alignment. After this point, similarity becomes more localized, indicating a transition toward more modality-specific representations in later layers.}
    \label{fig:crossmodal_similarity_esm3}
\end{wrapfigure}

\paragraph{Results.}
For ESM3, the cross-modal similarity matrix shows a clear two-phase structure (\cref{fig:crossmodal_similarity_esm3}). In the first 60\% of layers, sequence-only and structure-only representations are highly similar across a wide range of layer pairs, indicating that the two streams are strongly aligned early in depth. After this point, the similarity drops sharply, and the highest similarity becomes concentrated near a diagonal band in later layers, suggesting a phase transition where the two streams begin to diverge and develop more modality-specific computation. This transition closely matches the depth region where ESM3’s skiplayer effects and downstream gains become stronger in mid-to-late layers (\cref{fig:skiplayer_logitlens_structure_esm3,fig:skiplayer_layers_ESM3,fig:skiplayer_layers_multimodal_ESM3,fig:layerwise_average_spearman_ESM3,fig:tunedlens_ESM3}), and offers a plausible explanation for ESM3’s distinct behavior: early layers may primarily perform modality alignment, while later layers carry out the main task-relevant computation and refinement.

In contrast, DPLM2 does not exhibit the same pronounced early alignment-to-divergence transition (\cref{fig:crossmodal_similarity_dplm2}). Instead, the cross-modal similarity is more uniform across depth and becomes especially high for the largest variant, suggesting that sequence and structure representations remain more consistently coupled throughout the network. One possible reason for this behavior is that DPLM2 is initialized from a pretrained sequence-only DPLM and further trained on multimodal data. Furthermore, unlike ESM3, which has a separate tokenizer and prediction head for each modality, DPLM2 uses a shared tokenizer and prediction head across modalities, which may encourage a more homogeneous shared representation space rather than a sharp shift from alignment to modality-specific computation.

\paragraph{Summary.}
Overall, the modality ablations show that our depth analysis findings are not limited to the sequence pathway. Repeating the skiplayer, layer-wise similarity, and LogitLens analyses for the structure-only and joint sequence-structure settings in DPLM2 reveals the same trend as in the sequence stream: early-to-middle layers account for most of the propagated influence, while later layers mainly provide incremental refinement. At the same time, ESM3 keeps its distinct allocation of computation across depth. Cross-modal layer-layer similarity analysis for ESM3 shows strong sequence-structure alignment in the first 60\% of layers, followed by a sharp transition, consistent with its weaker early-layer effects and stronger mid-to-late contributions; this supports the hypothesis that early depth is used primarily for modality alignment and later depth for the main computation.\looseness=-1

\section{Discussion}
\label{sec:discussion}

\paragraph{Implications for depth efficiency in PLMs.}
Our results suggest that depth inefficiency is not limited to autoregressive LLMs or next-token prediction. Across autoregressive, masked, diffusion-based, and multimodal PLMs, we observe a broadly similar pattern: only a subset of layers carries most of the downstream-relevant computation, while other layers mainly provide incremental refinement. This is consistent with recent findings in LLMs, where later layers often have reduced downstream influence and mainly refine the final prediction \citep{csordas2025language, sun2025curse, lad2024remarkable, gromov2024unreasonable}. At the same time, the exact allocation of computation depends on the model family and modality. For example, ESM3 shows weaker early-layer effects and stronger mid-to-late contributions, which our cross-modal analysis suggests may be related to early sequence-structure alignment. This indicates that depth inefficiency may be a general property of deep Transformer-based PLMs, but its precise form can depend on how objectives, modalities, and architectures organize computation across depth.

\paragraph{Toward more adaptive and depth-efficient protein models.}
Efforts to address depth inefficiency in LLMs often rely on making computation more adaptive or improving layer utilization, for example, through Chain-of-Thought prompting, depth-dynamic models, or methods that directly target deeper-layer learning \citep{chen2025inner, sun2025curse,wei2022chain}. However, applying these ideas to protein models is not necessarily trivial. Protein sequences lack a clear notion of intermediate ``reasoning steps'', and their functional behavior emerges from distributed, nonlinear evolutionary and biophysical constraints rather than symbolic logic. That said, several directions appear promising. Growing approaches, where models are progressively expanded during pretraining, and looping approaches, where layers or blocks are reused across multiple computational steps, could be tested as ways to improve depth utilization in PLMs \citep{karp2024landscape,saunshi2024inductive,kapl2025depth,kapl2026growing,dehghani2018universal}. Sparse architectures such as MoE models may also be relevant, as our ProGen3 results show weaker scale-dependent sharpening of depth inefficiency, consistent with recent work suggesting that sparsity can mitigate the curse of depth in LLMs \citep{muhtar2026does}. Finally, our measurements could guide practical adaptive-compute methods for protein models, such as early-exit or layer-skipping policies that reduce inference cost while preserving downstream performance.

\section{Conclusion}
\label{sec:conclusion}

In this work, we ask whether protein language models exhibit a depth inefficiency analogous to the ``curse of depth'' observed in autoregressive LLMs. We study 7 PLM families and 26 model variants across autoregressive, masked, diffusion-based, and multimodal settings. Using a unified suite of layer-wise analyses, including layer-skipping interventions, representation similarity, layer-wise output readouts, and ProteinGym evaluation, we find that downstream-relevant computation is often concentrated in a subset of layers, while the remaining layers mainly provide incremental refinement. This pattern appears across training objectives and also persists in structure-only and joint sequence-structure settings, suggesting that depth inefficiency is a common feature of modern PLMs.

Our results motivate future work on more depth-efficient protein models. In particular, growing and looping schemes, sparse architectures, and adaptive-compute methods may help improve layer utilization and reduce inference cost. More broadly, understanding how computation is distributed across depth may provide a useful basis for designing PLMs that use scale more effectively across sequence, structure, and multimodal settings.

\section*{Acknowledgements}
This work was partially supported by the Wallenberg AI, Autonomous Systems and Software Program (WASP), funded by the Knut and Alice Wallenberg Foundation, and by the Helmholtz Foundation Model Initiative, supported by the Helmholtz Association.

The computations were enabled by the Berzelius resource, provided by the Knut and Alice Wallenberg Foundation at the National Supercomputer Centre, and by the Gauss Centre for Supercomputing e.V. (www.gauss-centre.eu), which provided the required computing time through the John von Neumann Institute for Computing (NIC) on the GCS Supercomputer JUPITER | JUWELS \citep{JUWELS} at Jülich Supercomputing Centre (JSC).

\bibliography{references}
\bibliographystyle{abbrvnat}

\clearpage
\appendix

\crefalias{section}{appendix}
\crefalias{subsection}{appendix}    %
\crefalias{subsubsection}{appendix} %

\section{Experimental Details}
\label{app:experimental_details}

In this section, we explain the included PLMs and provide full details of the experiments used in our study.

\subsection{Models}
\label{app:models}

\paragraph{ESM2.} ESM2 \citep{lin2023esm2} is a Transformer encoder protein language model trained with a masked language modeling (MLM) objective. It is commonly used as a general-purpose protein representation model across many different tasks, and for zero-shot fitness prediction. ESM2 is released in multiple sizes, from 8M to 15B, and is trained on UniRef50 \citep{suzek2007uniref}.

\paragraph{ESM3.} ESM3 \citep{hayes2025esm3} is a large-scale multimodal protein foundation model designed to jointly model protein \emph{sequence}, \emph{structure}, and \emph{function}. It is trained with an MLM objective and is primarily used for multimodal generation and design-style tasks, where one can condition on partial information of any of the modalities and generate consistent outputs. The ESM3 family is released at multiple scales of 1.4B, 7B, and 98B parameters, but only the smaller version is publicly available and is therefore used in our study.

\paragraph{Profluent-E1.} Profluent-E1 \citep{jain2025e1} is a family of protein Transformer encoder models trained with an MLM-style objective and designed for strong protein sequence representation learning. A key feature of E1 is that it can incorporate retrieved evolutionary context: homologous sequences are provided as additional inputs, and the model integrates this multi-sequence context through a dedicated attention scheme. E1 is released in three sizes, including 150M, 300M, and 600M parameters, and is trained for a large token budget on Profluent’s Protein Atlas \citep{bhatnagar2025progen3} and UniRef \citep{suzek2007uniref}.

\paragraph{DPLM.} DPLM \citep{wang2024dplm} is a diffusion-based protein language model trained under a discrete denoising diffusion objective. Rather than predicting masked tokens in one shot (MLM) or generating left-to-right (AR), DPLM learns to iteratively denoise corrupted sequences across diffusion timesteps while using bidirectional context. This makes DPLM useful both as a generative model via iterative sampling and as a representation learner that can be fine-tuned or used zero-shot for downstream prediction tasks. DPLM is released in multiple sizes, including 150M, 650M, and 3B, and is trained on UniRef50 \citep{suzek2007uniref}.

\paragraph{DPLM2.} DPLM2 \citep{wang2025dplm2} extends DPLM to a multimodal diffusion model that jointly models protein \emph{sequence and 3D structure}. It converts 3D coordinates into discrete structure tokens and trains a unified denoising model over the combined token space, enabling tasks such as folding, inverse folding, and multimodal scaffolding through conditional diffusion. It is released in the same sizes as DPLM and is initialized from a pretrained DPLM before being further trained on a structure-supervised multimodal dataset built from experimental and high-quality synthetic structures.

\paragraph{ProGen2.} ProGen2 \citep{nijkamp2023progen2} is an autoregressive Transformer decoder model trained with next-token prediction. It is mainly used for protein sequence generation and for zero-shot fitness prediction. ProGen2 is released in multiple sizes, including small, medium, base, large, and xlarge models that scale up to 6.4B parameters, and is trained on a curated large-scale protein sequence corpus. Given that ProGen2-base has the same number of parameters as the medium variant, we exclude it from our study.

\paragraph{ProGen3.} ProGen3 \citep{bhatnagar2025progen3} is an autoregressive protein language model family based on a sparse \emph{Mixture-of-Experts (MoE)} Transformer decoder, trained with next-token prediction. It is designed for scalable protein sequence generation, where sparsity enables scaling model capacity while keeping per-token computation manageable. ProGen3 is trained on Profluent’s large curated protein sequence corpus (\emph{Protein Atlas}) \citep{bhatnagar2025progen3}, and is released publicly in 6 model scales, from 112M to 3B parameters.

\subsection{Evaluation Setup}
\label{app:evaluation_setup}

\subsubsection{Propagated Layer Skipping and Similarity Analysis}
\label{app:skiplayer_details}

We follow the layer-skipping tools of \citet{csordas2025language} to measure how much each layer contributes to downstream computation by removing that layer's residual update and tracking how the change propagates through later layers. For a Transformer with residual stream states $h_\ell$ at layer $\ell$, we view each layer as applying an additive update $\Delta h_\ell = h_{\ell+1} - h_\ell$. Skipping a source layer $s$ is implemented by removing its update, i.e., setting $\bar h_{s+1}:= \bar h_s$, and then continuing the forward pass normally to obtain intervened activations $\{\bar h_\ell\}_{\ell > s}$. We then quantify the propagated effect of skipping $s$ on a later layer $\ell$ by comparing the downstream representations under the original and intervened forward passes. In practice, we aggregate these changes across token positions and prompts using a simple summary statistic of the maximum effect over relevant token positions and prompts, to produce the layer-by-layer heatmaps. Additionally, we also report the layer-layer similarity across depth for all models. We measure the similarity between a pair of layers using the linear centered kernel alignment (CKA) \citep{kornblith2019cka}.

To isolate importance for future predictions, we use the \textit{skiplayer future} variant. For autoregressive models (ProGen2 and ProGen3), we sample a split position $1 < t_s < T-1$, apply the layer-skip intervention only to token positions $t \le t_s$, and then measure the effect only on positions $t > t_s$. This directly tests whether computations performed for earlier tokens are reused when predicting later tokens. For masked and diffusion-based models that do not have a left-to-right notion of ``future'', we adapt \textit{skiplayer future} to capture cross-positional effects by replacing future tokens with held-out, non-intervened positions: we randomly sample an intervention set of positions and apply the skip only on that subset, and then measure propagated effects on the remaining non-intervened positions. Concretely, for each prompt, we randomly sample between 20\% and 80\% of masked positions and between 20\% and 80\% of non-masked positions for intervention, and compute effects on the remaining masked positions. We repeat this 4 times per prompt and report the maximum effects. We additionally measure effects on all non-intervened tokens (i.e., both masked and non-masked) and observe that both choices yield similar trends. Hence, we report results only for the former, as predictions for masked token positions are most sensitive to perturbations.
In addition to representation-level effects, we also quantify how skipping each layer changes the model's final outputs in the \textit{skiplayer future} setting. For each skipped layer, we compute the $\ell_2$ change in the output vector between the original and intervened runs, i.e., $\|y - \bar y\|_2$, and report the maximum change over the evaluated future positions. We use this norm-based metric rather than KL divergence because it yields clearer, more stable visualizations for this intervention setting. For all experiments, we use 40 protein sequences randomly sampled from UniRef50 \citep{suzek2007uniref} as input prompts.

\subsubsection{Layer-wise output readouts with TunedLens and LogitLens}
\label{app:lens_details}

To track how the model's implied prediction distribution evolves across depth, we use TunedLens and LogitLens readouts \citep{belrose2023eliciting,logitlens} applied to intermediate residual representations. After each layer $\ell$, we take the residual stream at that depth and map it directly to token logits using an unembedding head. In LogitLens, we directly apply the model's standard unembedding, i.e., the final normalization followed by the output projection, to the intermediate representations. In TunedLens, we first train a linear adapter per layer using a subset of UniRef50 dataset, to improve the alignment between intermediate representations and the final layer output space. At inference, we pass the intermediate representations per layer through the corresponding pretrained adapter before applying the model's unembedding module. To train the per-layer adapters, we use the default setup as in the original TunedLens project \citep{belrose2023eliciting}, with a few changes to suit our setting - one, we use a subset of 435,458 protein sequences randomly sampled from UniRef50 (with no overlap to the 40 sequences used in our analysis) as the training data; two, for masked and diffusion models, we train only on the masked tokens; and finally, since we handle models across different scales, we set the training steps to be dependent on the model's hidden dimension: for models with hidden size below 1000, we use 250 steps, between 1000 and 2000 we use 500 steps, and beyond that we train for up to 750 steps. We report TunedLens results for all sequence-stream analyses across models, and LogitLens results for the structure- and multimodal-stream analyses of ESM3 and DPLM2. After obtaining the logits per layer, we apply softmax and retrieve a layer-wise distribution $p_\ell(\cdot)$ over the vocabulary. We compare this intermediate distribution to the final distribution $p_L(\cdot)$ produced by the full model at the last layer $L$.

We report two complementary summary metrics. First, we compute the KL divergence $D_{\mathrm{KL}}(p_L \,\|\, p_\ell)$ to measure how close the layer-wise distribution is to the final prediction. Second, we measure discrete agreement with the final prediction using top-$k$ overlap. Since protein vocabularies are much smaller than natural-language vocabularies, essentially limited to the amino-acid alphabet, we report top-1 overlap ($k=1$): whether the most likely token under $p_\ell$ matches the most likely token under $p_L$. We aggregate these metrics over token positions and prompts on the same UniRef50 prompt set as in the previous experiments and report the results. In the case of masked and diffusion-based models, we aggregate only over the masked tokens.

\subsubsection{Layer-wise ProteinGym Evaluation}
\label{app:proteingym_details}

To connect previous depth measurements to practical utility, we evaluate layer-wise performance on ProteinGym \citep{notin2023proteingym}, a large benchmark of mutation effect prediction with 217 standardized deep mutational scanning (DMS) substitution assays. For each assay, the goal is to assign a score to each variant sequence and report Spearman's rank correlation between predicted scores and experimental measurements. We follow the official ProteinGym evaluation protocol and codebase.

For each model, we compute a mutant--wild-type score using likelihood-based scoring as implemented in the official repository, with a layer-wise early-exit variant for our depth analysis. Concretely, for a chosen layer $\ell$, we obtain logits from that layer via early exit and compute mutation scores exactly as in the full model, but using the layer-$\ell$ logits. For masked- and diffusion-based models such as ESM2, ESM3, DPLM, DPLM2, and Profluent-E1, we use masked-marginals scoring \citep{meier2021language}: for each mutated position, we mask that position, compute the log-probability of the mutant amino acid and the wild-type amino acid at that position, and take their difference. For multi-mutation variants, we take the sum of per-mutation differences. For ProGen2 and ProGen3, we use the ProteinGym repository's autoregressive scoring procedure. We then report Spearman correlation per assay and summarize performance by averaging across all assays.

To make results comparable across model sizes with different numbers of layers, we plot performance against relative depth, normalizing layer index to $[0,1]$ within each model. We also follow ProteinGym's standard handling of long sequences when needed so that scores are computed consistently under each model's maximum context length. Overall, this evaluation provides a downstream counterpart to the mechanistic measurements: it measures how quickly a useful predictive signal emerges with depth and where performance gains saturate when increasing the depth.

\section{Additional Results}
\label{app:additional_results}

Here we provide additional results for the experiments in \cref{sec:experiments}. Figures are grouped by model family and shown across publicly available sizes. For skiplayer experiments, heatmaps report the propagated effect of skipping a source layer on later-layer representations, measured on future tokens, alongside layer-layer CKA similarity. Additionally, the output-norm plots summarize the corresponding change in final outputs. For TunedLens and LogitLens experiments, we report KL divergence and top-1 overlap between layer-wise predictions and the full-model prediction. For ProteinGym, we report the average Spearman correlation across assays as a function of relative depth.

\subsection{How much does each layer contribute to later computations?}
\label{app:skiplayer_effects}

\begin{figure}[h]
    \centering
    \includegraphics[width=0.9\textwidth]{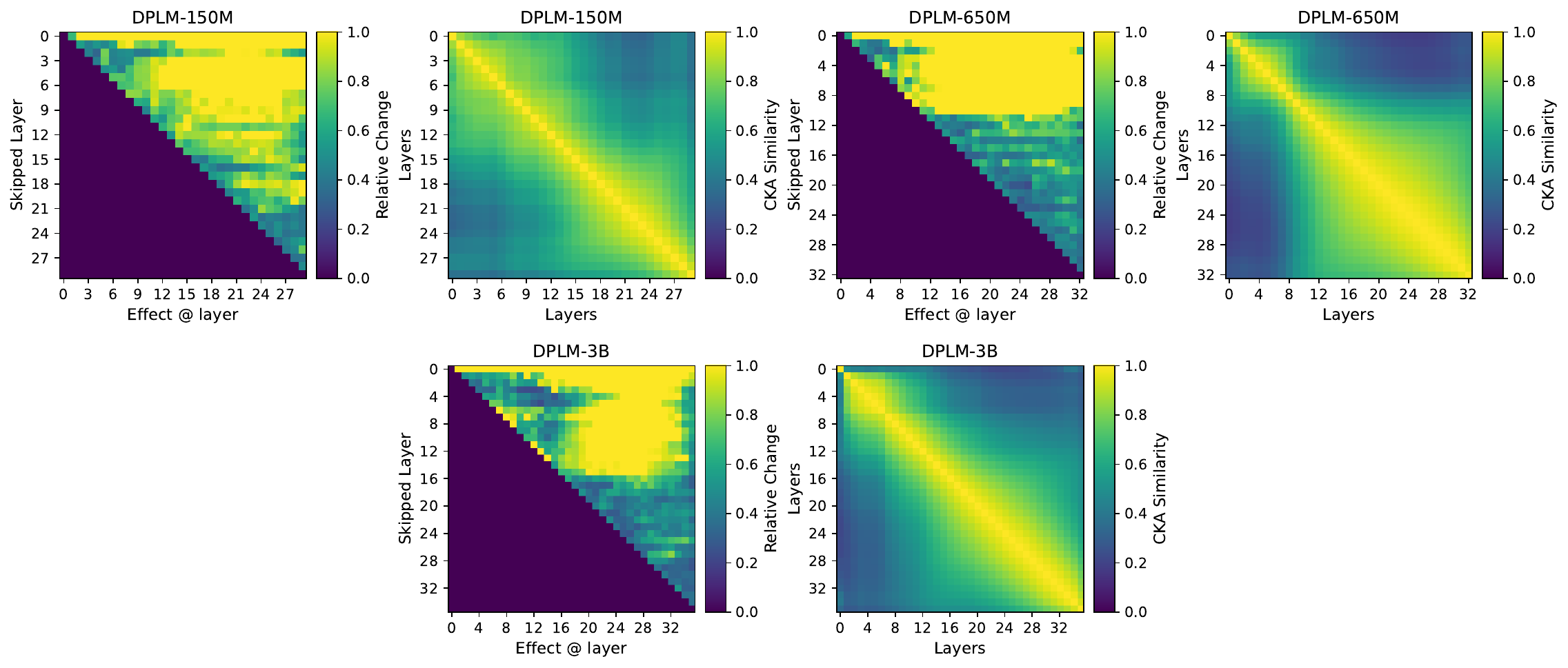}
    \caption{Maximum propagated effect of skipping each layer on future-token computations for DPLM.}
    \label{fig:skiplayer_layers_DPLM}
\end{figure}

\begin{figure}[h]
    \centering
    \includegraphics[width=0.9\textwidth]{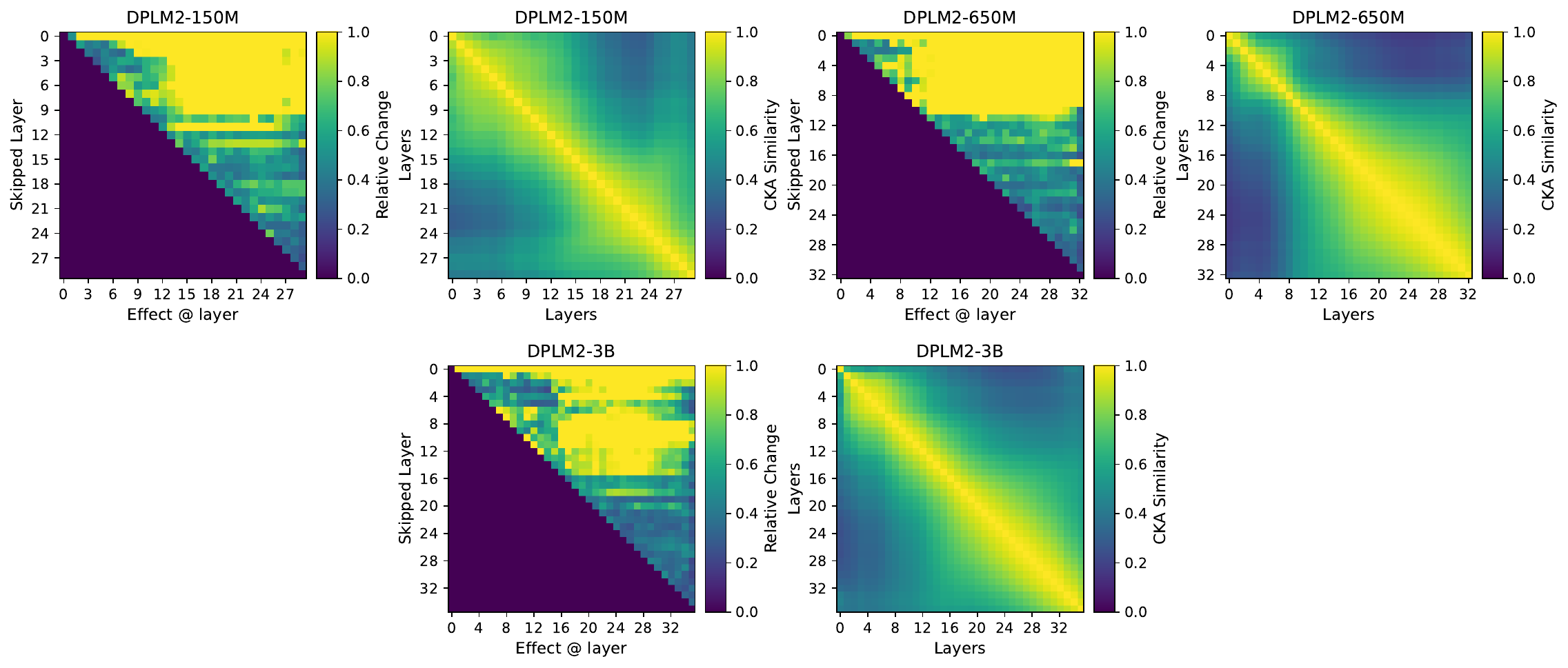}
    \caption{Maximum propagated effect of skipping each layer on future-token computations for DPLM2.}
    \label{fig:skiplayer_layers_DPLM2}
\end{figure}

\begin{figure}[h]
    \centering
    \includegraphics[width=0.9\textwidth]{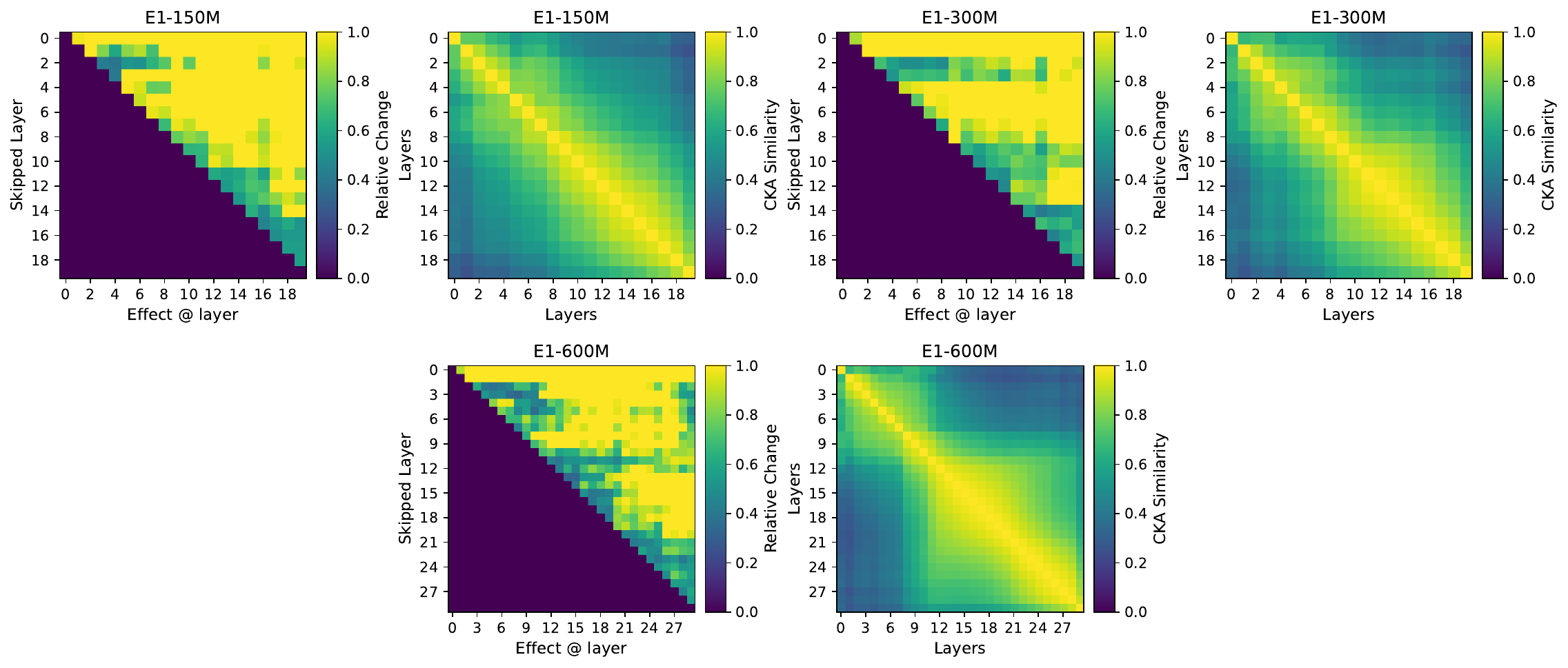}
    \caption{Maximum propagated effect of skipping each layer on future-token computations for Profluent-E1.}
    \label{fig:skiplayer_layers_E1}
\end{figure}

\begin{figure}[h]
    \centering
    \includegraphics[width=0.9\linewidth]{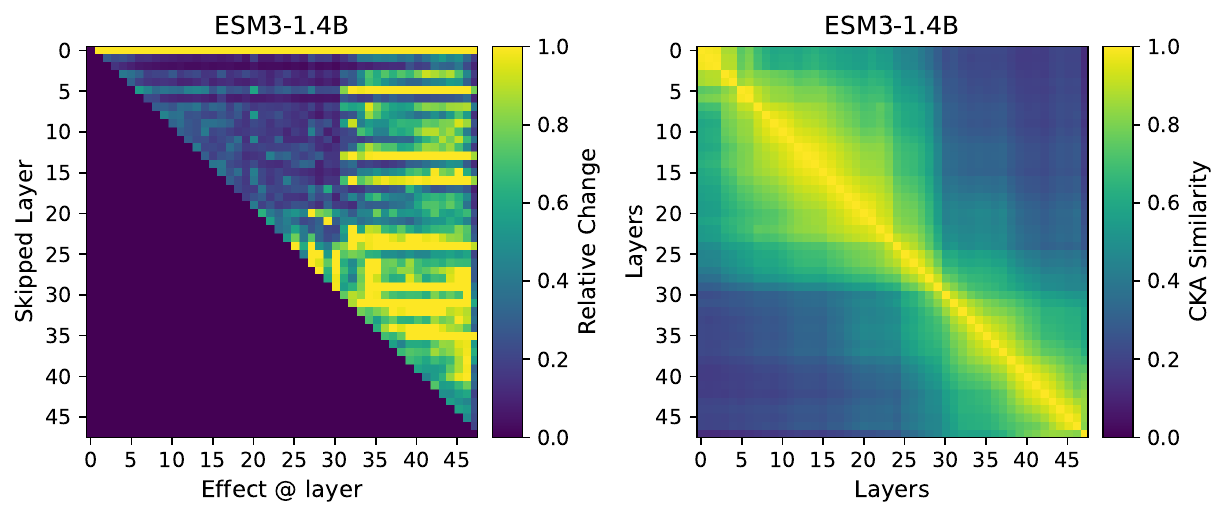}
    \caption{Maximum propagated effect of skipping each layer on later-layer representations (future tokens only), for ESM3.}
    \label{fig:skiplayer_layers_ESM3}
\end{figure}

\begin{figure}[h]
    \centering
    \includegraphics[width=0.9\textwidth]{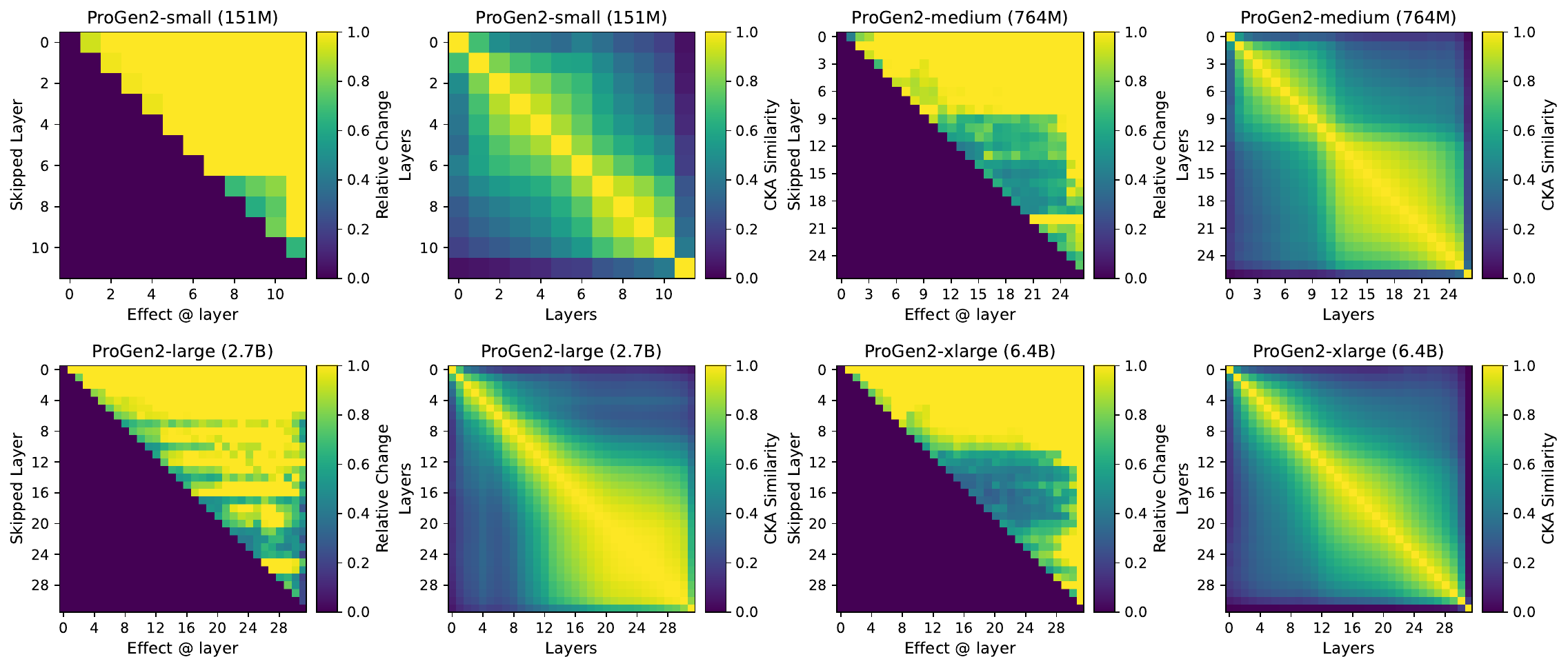}
    \caption{Maximum propagated effect of skipping each layer on future-token computations for ProGen2.}
    \label{fig:skiplayer_layers_ProGen2}
\end{figure}

\begin{figure}[h]
    \centering
    \includegraphics[width=0.9\textwidth]{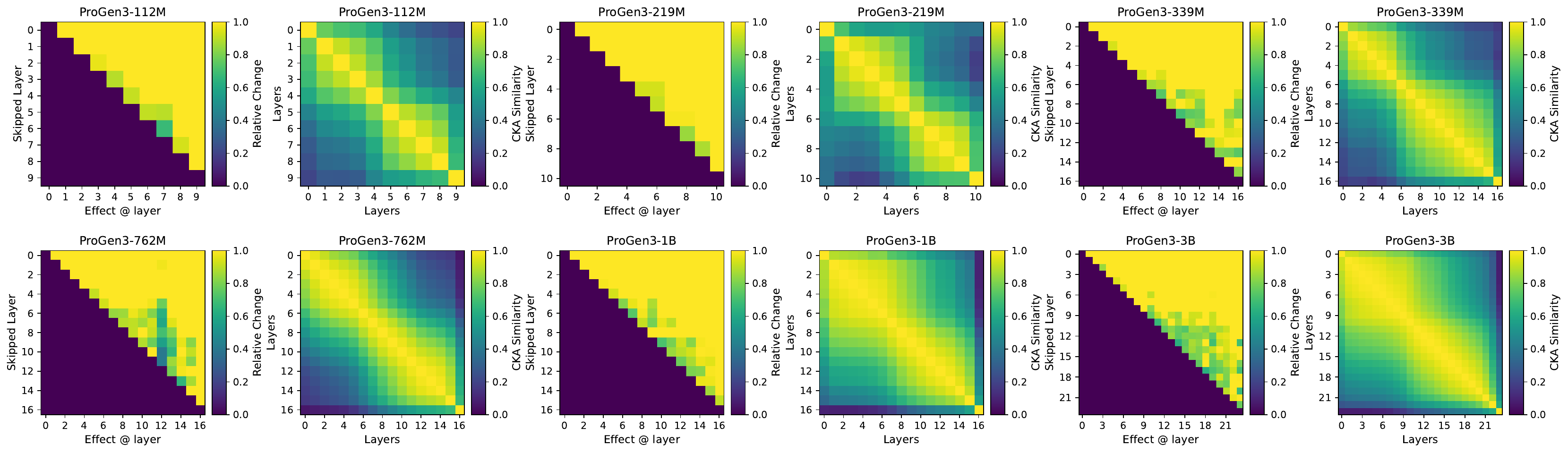}
    \caption{Maximum propagated effect of skipping each layer on future-token computations for ProGen3.}
    \label{fig:skiplayer_layers_ProGen3}
\end{figure}

\cref{fig:skiplayer_layers_ESM3,fig:skiplayer_layers_E1,fig:skiplayer_layers_DPLM,fig:skiplayer_layers_DPLM2,fig:skiplayer_layers_ProGen2,fig:skiplayer_layers_ProGen3} report propagated \textit{skiplayer future} effects for ESM3, Profluent-E1, DPLM, DPLM2, ProGen2 and ProGen3, respectively. Recall that \textit{skiplayer future} isolates whether computations performed for earlier tokens are reused when predicting held-out (non-intervened) positions. We quantify propagation at the representation level as the maximum change in downstream-layer hidden states on the future token positions when skipping each source layer. We also report layer-layer similarity alongside the effects heatmaps. Across model families, propagated effects are typically strongest when skipping early-to-middle layers and weaker for many late layers, with the separation often becoming clearer at larger scales. We also observe localized low-effect regions in early layers in some models, which supports the stage-like structure of the network across depth \citep{lad2024remarkable}.

\begin{figure}[h]
    \centering
    \includegraphics[width=0.7\textwidth]{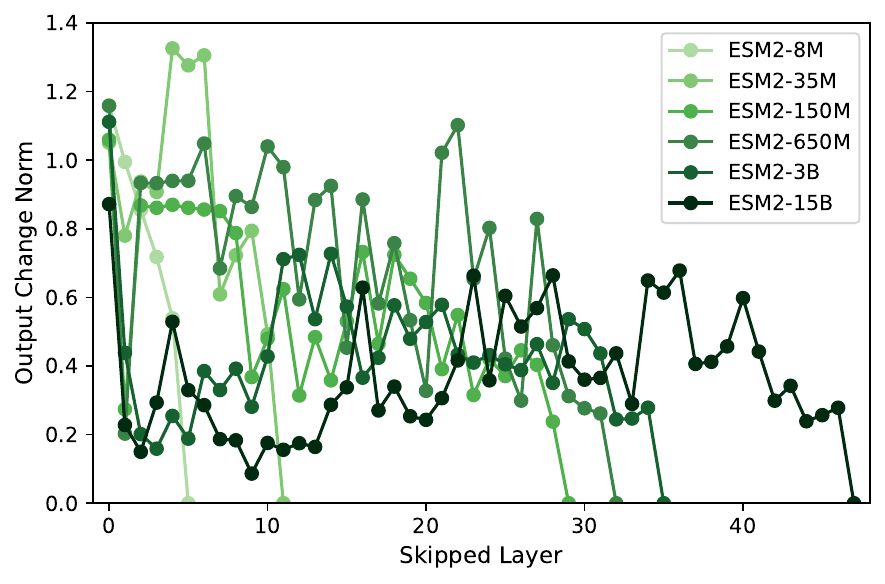}
    \caption{
    Maximum change in ESM2 output probabilities under layer skipping, restricted to future tokens only.}
    \label{fig:skiplayer_outnorm_ESM2}
\end{figure}

\begin{figure}[h]
    \centering
    \includegraphics[width=0.7\textwidth]{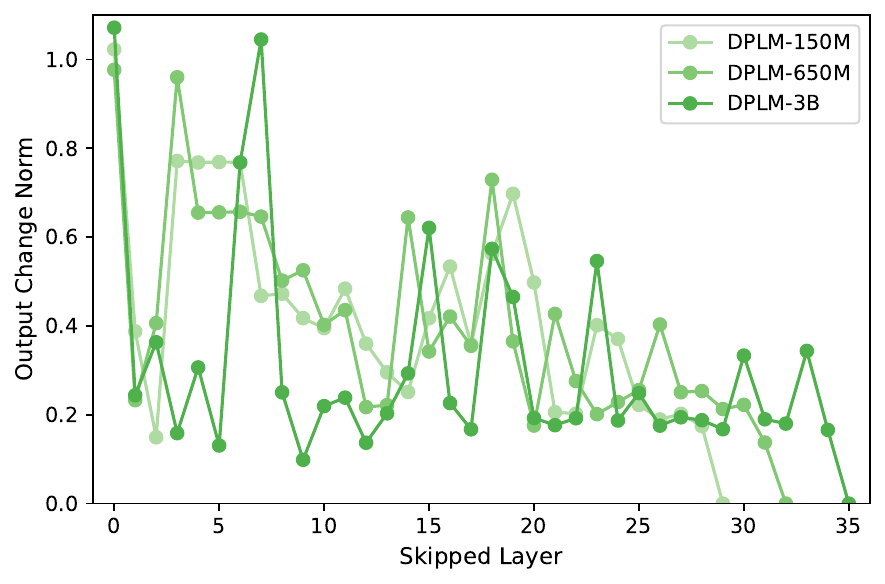}
    \caption{Maximum change in DPLM output probabilities under layer skipping, restricted to future tokens only.}
    \label{fig:skiplayer_outnorm_DPLM}
\end{figure}

\begin{figure}[h]
    \centering
    \includegraphics[width=0.7\textwidth]{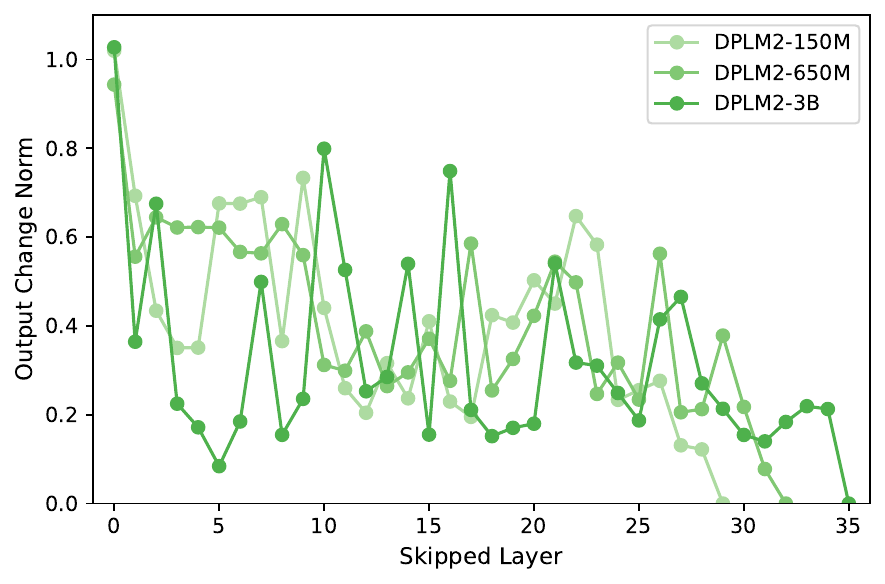}
    \caption{Maximum change in DPLM2 output probabilities under layer skipping, restricted to future tokens only.}
    \label{fig:skiplayer_outnorm_DPLM2}
\end{figure}

\begin{figure}[h]
    \centering
    \includegraphics[width=0.7\textwidth]{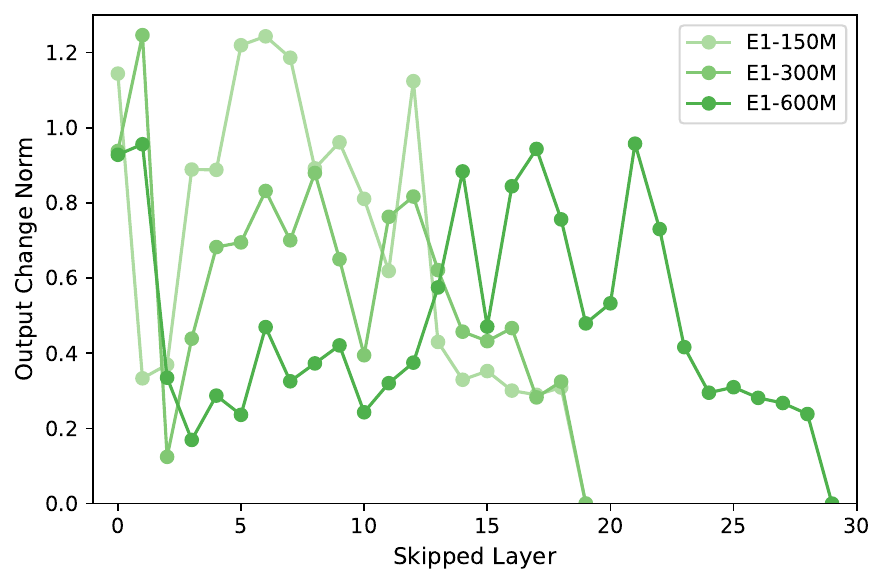}
    \caption{Maximum change in Profluent-E1 output probabilities under layer skipping, restricted to future tokens only.}
    \label{fig:skiplayer_outnorm_E1}
\end{figure}

\begin{figure}[h]
    \centering
    \includegraphics[width=0.7\linewidth]{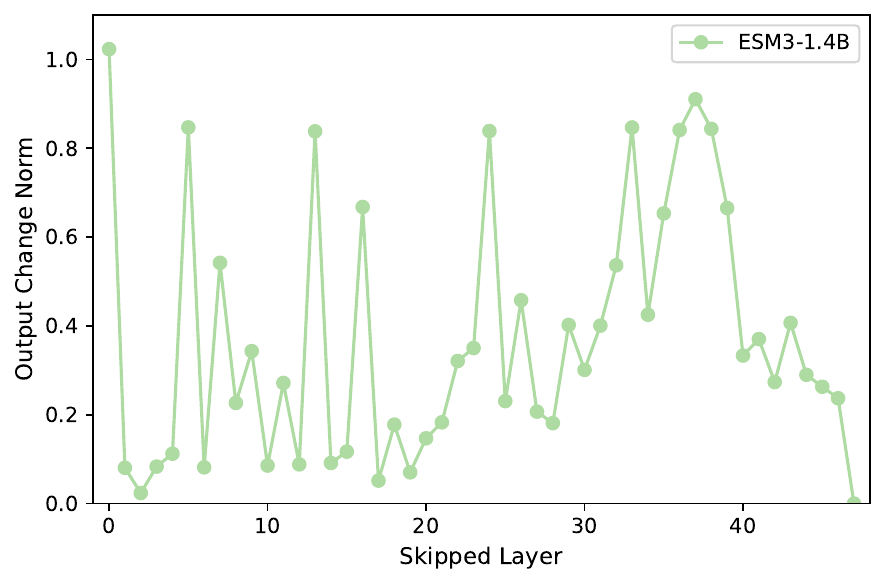}
    \caption{Maximum change in ESM3 output probabilities under layer skipping, restricted to future tokens only.}
    \label{fig:skiplayer_outnorm_ESM3}
\end{figure}

\begin{figure}[h]
    \centering
    \includegraphics[width=0.7\textwidth]{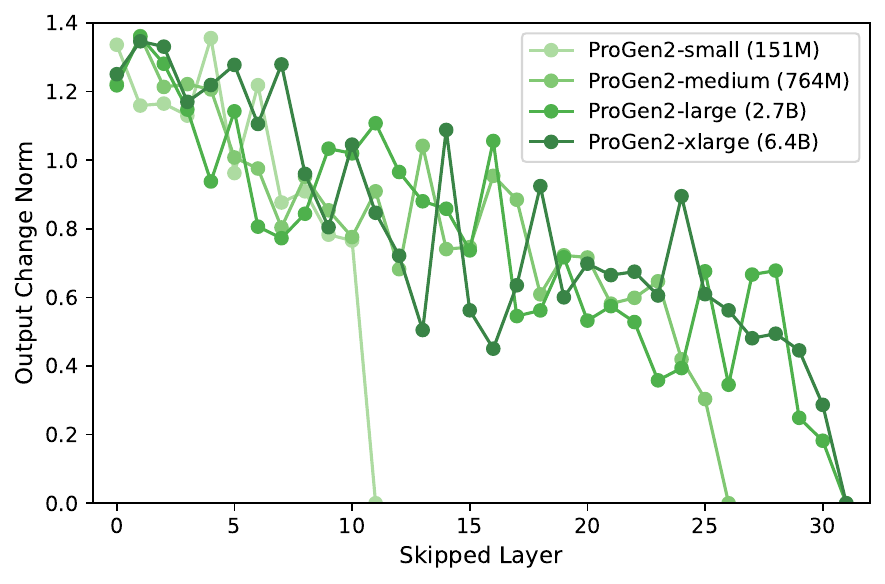}
    \caption{Maximum change in ProGen2 output probabilities under layer skipping, restricted to future tokens only.}
    \label{fig:skiplayer_outnorm_ProGen2}
\end{figure}

\begin{figure}[h]
    \centering
    \includegraphics[width=0.7\textwidth]{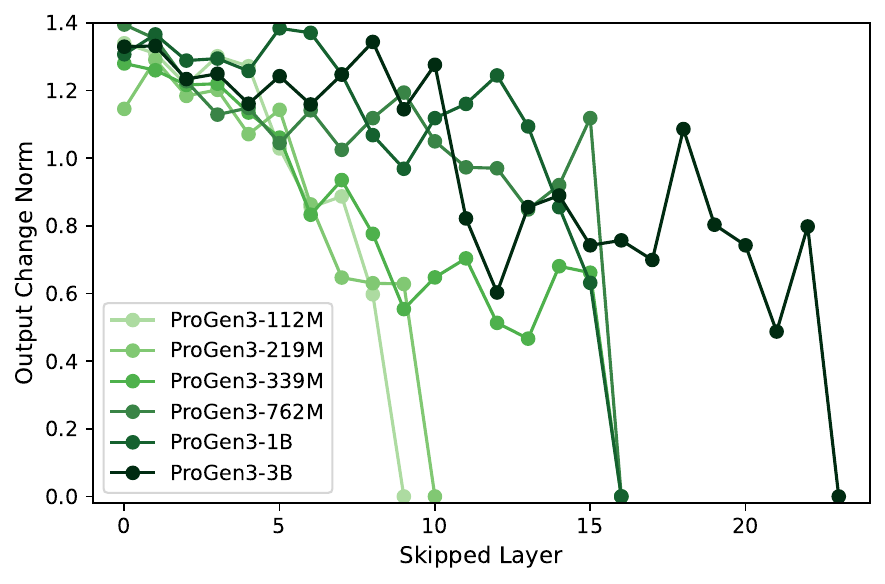}
    \caption{Maximum change in ProGen3 output probabilities under layer skipping, restricted to future tokens only.}
    \label{fig:skiplayer_outnorm_ProGen3}
\end{figure}

\cref{fig:skiplayer_outnorm_ESM2,fig:skiplayer_outnorm_ESM3,fig:skiplayer_outnorm_E1,fig:skiplayer_outnorm_DPLM,fig:skiplayer_outnorm_DPLM2,fig:skiplayer_outnorm_ProGen2,fig:skiplayer_outnorm_ProGen3} report the corresponding output-level sensitivity under the same setup, measured as the maximum change in the model outputs. While representation-level propagation captures how strongly information flows forward through the network, output-level sensitivity reflects how much this propagated change ultimately affects the model's predictions. Sharp peaks indicate layers whose computations are strongly reused for future predictions, whereas flatter profiles suggest weaker reuse. For ESM2 (\cref{fig:skiplayer_outnorm_ESM2}), we observe an overall downward trend with depth, with small fluctuations and minor differences across model sizes. ESM3 (\cref{fig:skiplayer_outnorm_ESM3}) shows the opposite pattern: early layers have the weakest effect on future outputs, while the most impactful layers cluster around mid-depth, which is consistent with its skiplayer patterns in \cref{fig:skiplayer_layers_ESM3}. For DPLM, DPLM2, and Profluent-E1, we find a more consistent decrease with depth, where deeper layers tend to induce smaller changes in future outputs (\cref{fig:skiplayer_outnorm_E1,fig:skiplayer_outnorm_DPLM,fig:skiplayer_outnorm_DPLM2}). This effect is particularly pronounced in ProGen2 and ProGen3 (\cref{fig:skiplayer_outnorm_ProGen2,fig:skiplayer_outnorm_ProGen3}), where output-norm differences consistently decrease with depth across all sizes.

\subsection{How does the probability distribution vary across layers?}
\label{app:lens}

\begin{figure}[h]
    \centering
    \includegraphics[width=0.9\linewidth]{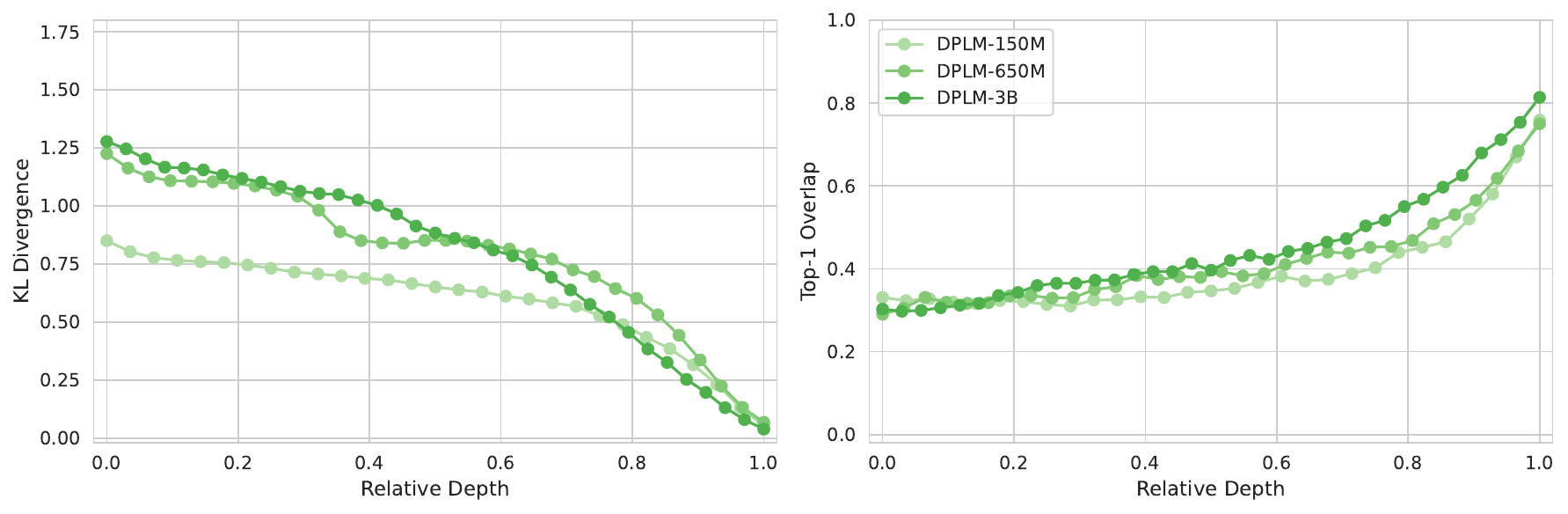}
    \caption{TunedLens analysis for DPLM across depth: KL divergence between the layer-wise output distribution and the final output distribution (left), and top-1 overlap between the layer-wise prediction and the full-model prediction (right).}
    \label{fig:tunedlens_DPLM}
\end{figure}

\begin{figure}[h]
    \centering
    \includegraphics[width=0.9\linewidth]{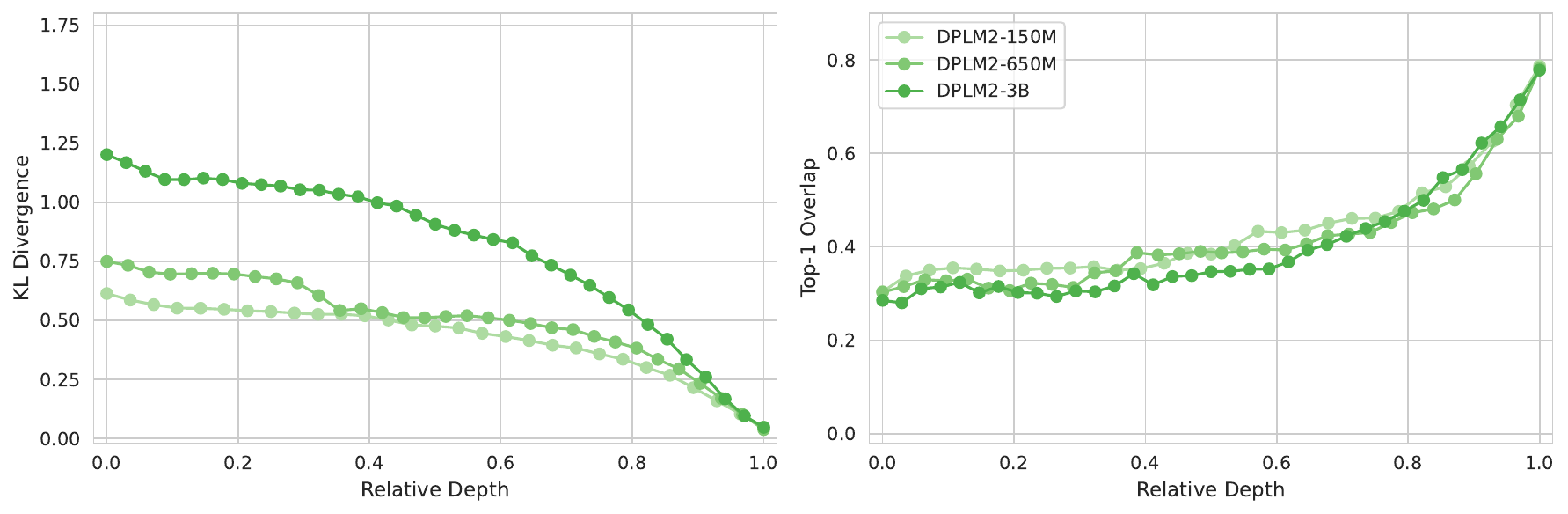}
    \caption{TunedLens analysis for DPLM2 across depth: KL divergence between the layer-wise output distribution and the final output distribution (left), and top-1 overlap between the layer-wise prediction and the full-model prediction (right).}
    \label{fig:tunedlens_DPLM2}
\end{figure}

\begin{figure}[h]
    \centering
    \includegraphics[width=0.9\linewidth]{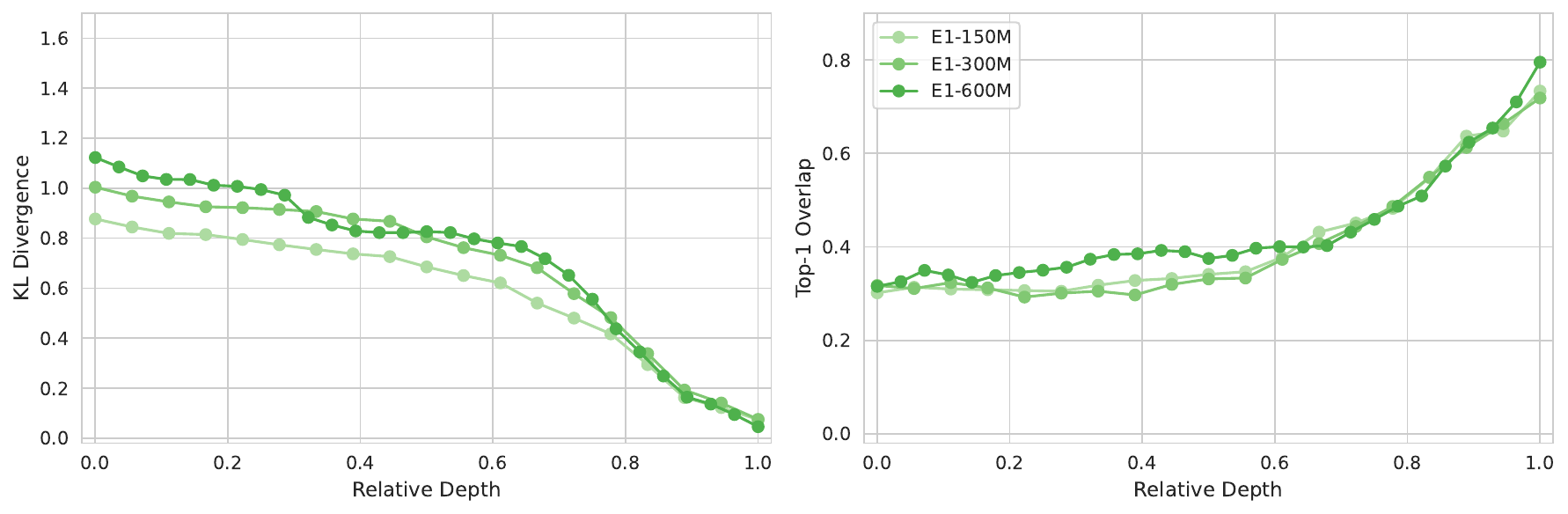}
    \caption{TunedLens analysis for E1 across depth: KL divergence between the layer-wise output distribution and the final output distribution (left), and top-1 overlap between the layer-wise prediction and the full-model prediction (right).}
    \label{fig:tunedlens_E1}
\end{figure}

\begin{figure}[h]
    \centering
    \includegraphics[width=0.9\linewidth]{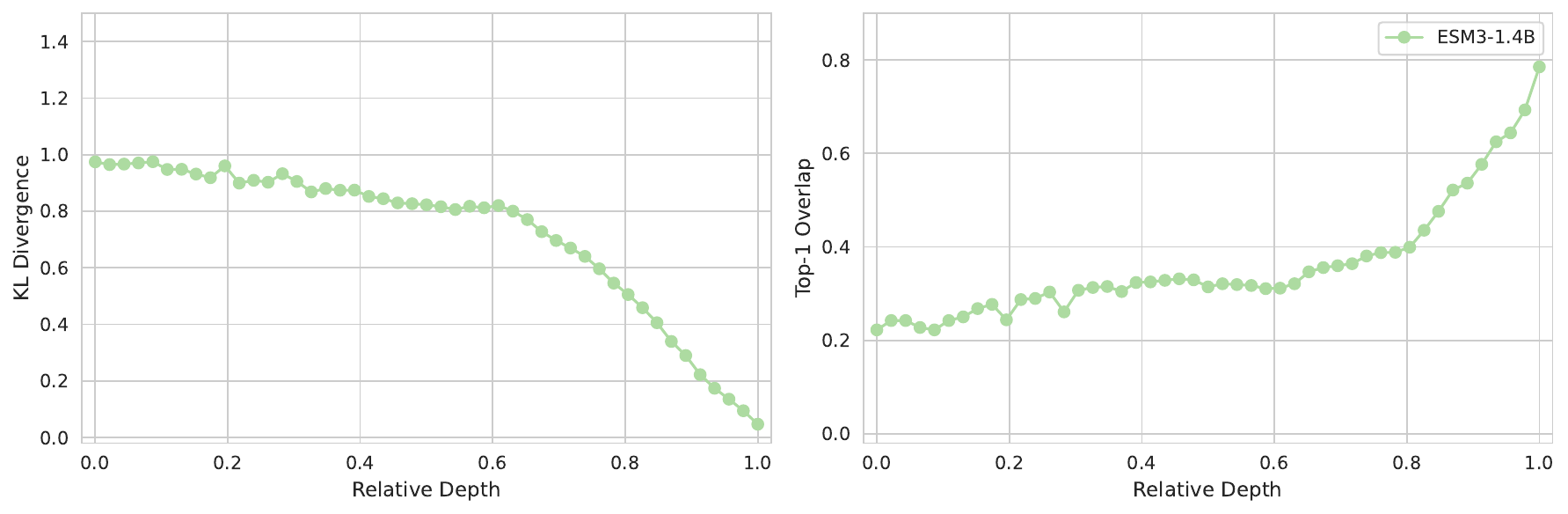}
    \caption{TunedLens analysis for ESM3 across depth: KL divergence between the layer-wise output distribution and the final output distribution (left), and top-1 overlap between the layer-wise prediction and the full-model prediction (right).}
    \label{fig:tunedlens_ESM3}
\end{figure}

\begin{figure}[h]
    \centering
    \includegraphics[width=0.9\linewidth]{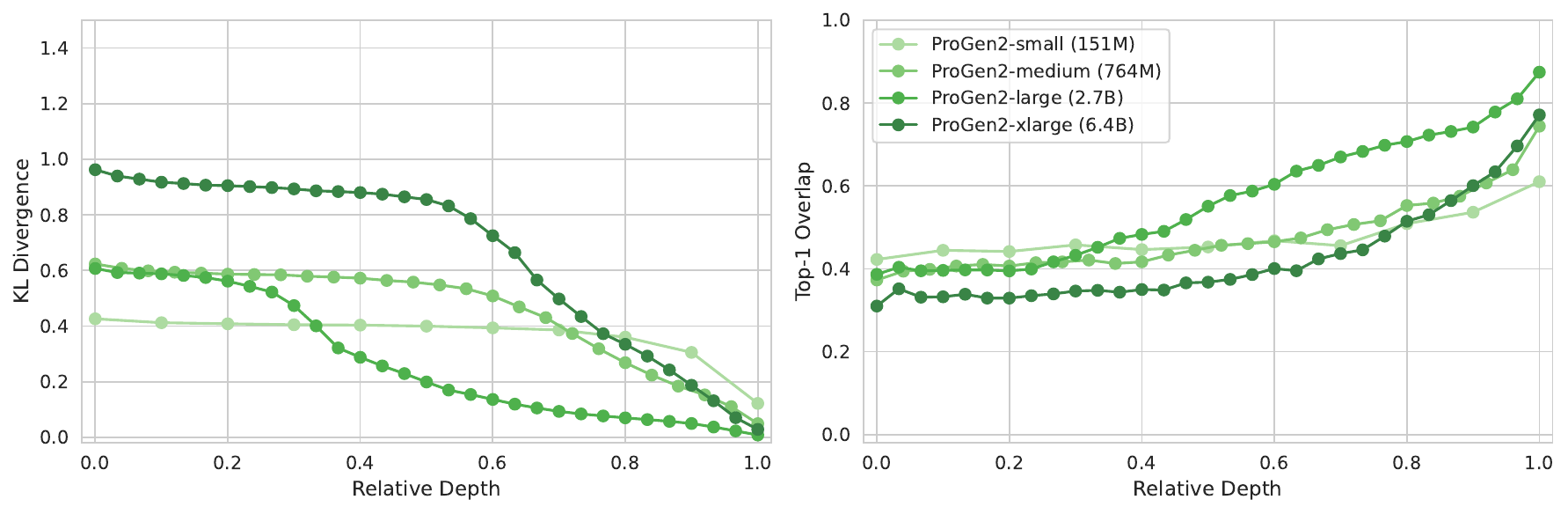}
    \caption{TunedLens analysis for ProGen2 across depth: KL divergence between the layer-wise output distribution and the final output distribution (left), and top-1 overlap between the layer-wise prediction and the full-model prediction (right).}
    \label{fig:tunedlens_ProGen2}
\end{figure}

\begin{figure}[h]
    \centering
    \includegraphics[width=0.9\linewidth]{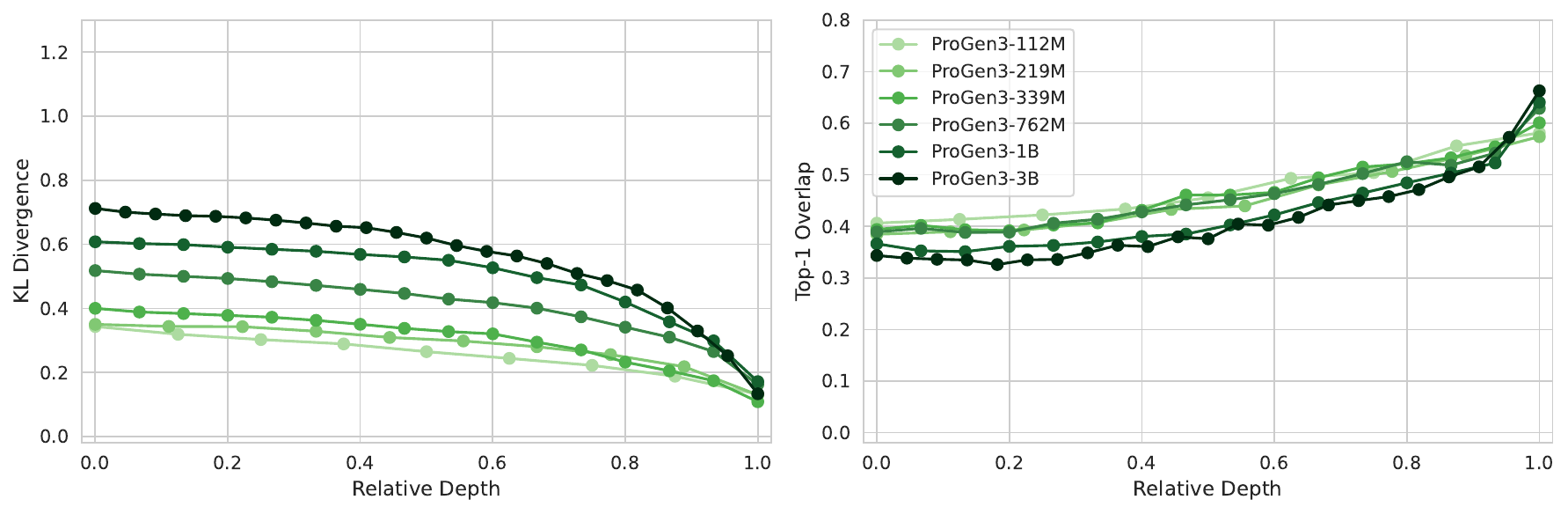}
    \caption{TunedLens analysis for ProGen3 across depth: KL divergence between the layer-wise output distribution and the final output distribution (left), and top-1 overlap between the layer-wise prediction and the full-model prediction (right).}
    \label{fig:tunedlens_ProGen3}
\end{figure}

\cref{fig:tunedlens_ESM3,fig:tunedlens_E1,fig:tunedlens_DPLM,fig:tunedlens_DPLM2,fig:tunedlens_ProGen2,fig:tunedlens_ProGen3} report how the model’s token-level output distribution evolves across depth, using a layer-wise readout and comparing each layer’s distribution to the final-layer distribution. Lower KL divergence indicates that a layer already produces a distribution close to the final model output, whereas higher KL divergence suggests that substantial refinement still occurs in later layers.

Complementing this distribution-level view, the same figures report the top-1 overlap between the layer-wise argmax prediction and the final model’s argmax prediction across depth. Top-1 overlap is easier to interpret but can hide meaningful distributional changes when probability mass shifts without changing the argmax, so we report both metrics together. Overall, these plots help distinguish whether later layers mainly sharpen already-formed predictions, i.e., low KL and high top-1 early, or whether they continue to meaningfully change the predicted distribution, by showing persistently high KL and low top-1.

Overall, across all models, KL divergence generally decreases with depth and top-1 overlap increases, indicating that layer-wise predictions progressively align with the full-model output. For the largest variants in some families, these curves can become flat earlier in depth, suggesting that the refinement phase begins relatively early in the network.

\newpage

\subsection{How do different layers affect downstream performance?}
\label{app:proteingym}

\begin{figure}[h]
    \centering
    \includegraphics[width=0.7\linewidth]{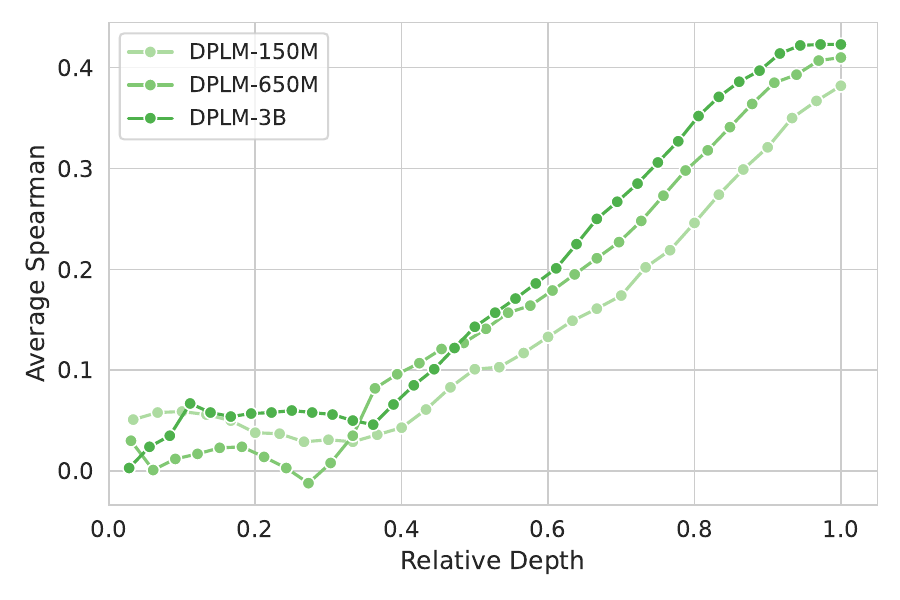}
    \caption{Average Spearman correlation for DPLM on ProteinGym, calculated at each layer. The relative depth is normalized to $[0,1]$.}
    \label{fig:layerwise_average_spearman_DPLM}
\end{figure}

\begin{figure}[h]
    \centering
    \includegraphics[width=0.7\linewidth]{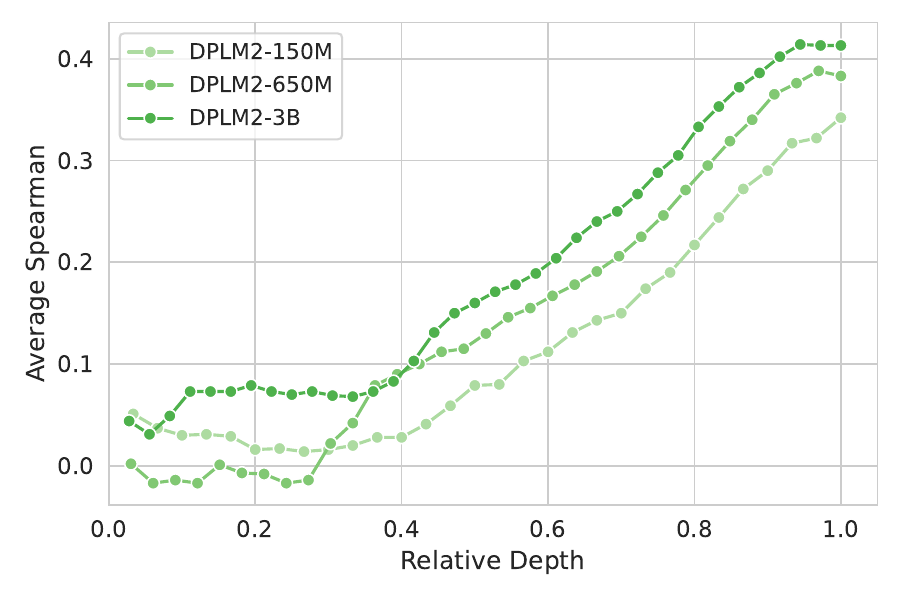}
    \caption{Average Spearman correlation for DPLM2 on ProteinGym, calculated at each layer. The relative depth is normalized to $[0,1]$.}
    \label{fig:layerwise_average_spearman_DPLM2}
\end{figure}

\begin{figure}[h]
    \centering
    \includegraphics[width=0.7\linewidth]{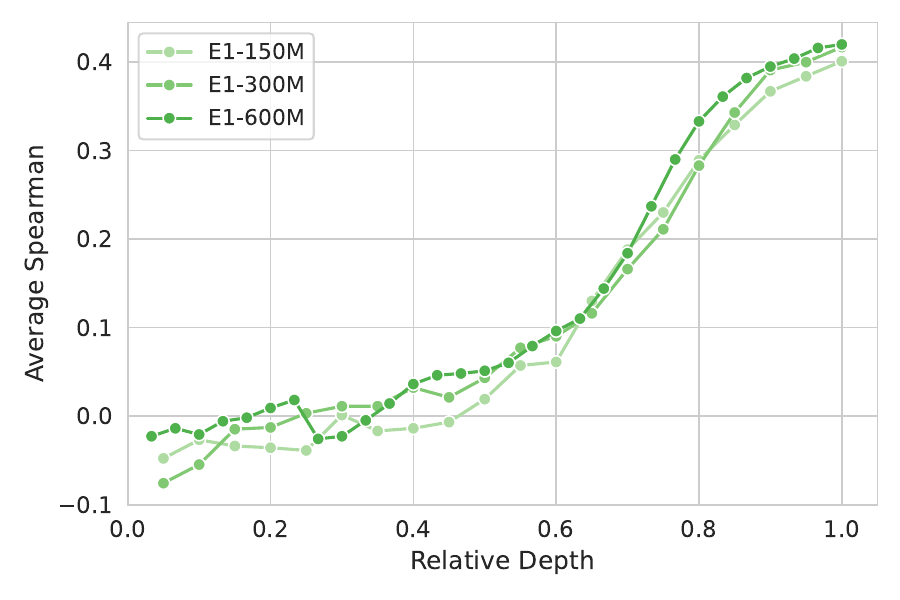}
    \caption{Average Spearman correlation for Profluent-E1 on ProteinGym, calculated at each layer. The relative depth is normalized to $[0,1]$.}
    \label{fig:layerwise_average_spearman_E1}
\end{figure}

\begin{figure}[h]
    \centering
    \includegraphics[width=0.7\linewidth]{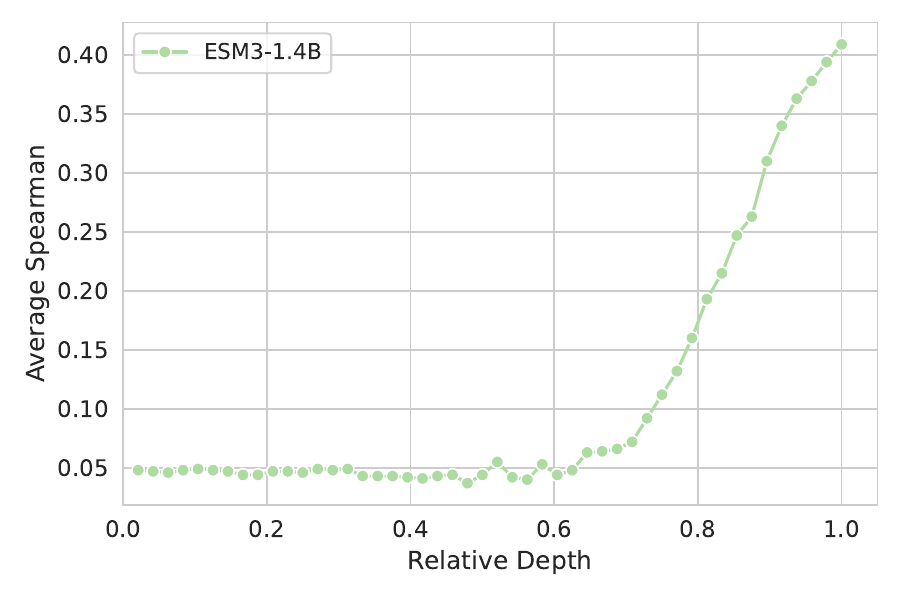}
    \caption{Average Spearman correlation for ESM3 on ProteinGym, calculated at each layer. The relative depth is normalized to $[0,1]$.}
    \label{fig:layerwise_average_spearman_ESM3}
\end{figure}

\begin{figure}[h]
    \centering
    \includegraphics[width=0.7\linewidth]{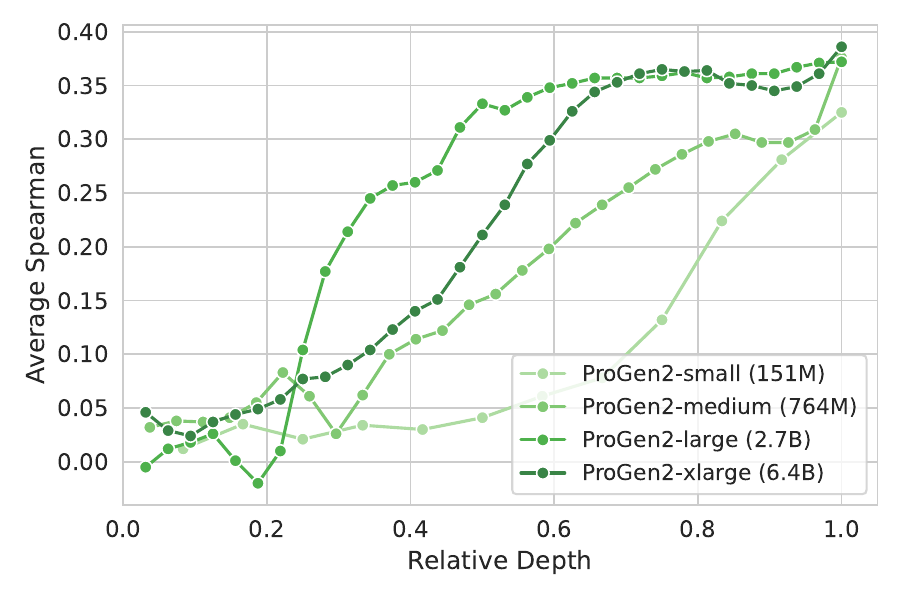}
    \caption{Average Spearman correlation for ProGen2 on ProteinGym, calculated at each layer. The relative depth is normalized to $[0,1]$.}
    \label{fig:layerwise_average_spearman_ProGen2}
\end{figure}

\begin{figure}[h]
    \centering
    \includegraphics[width=0.7\linewidth]{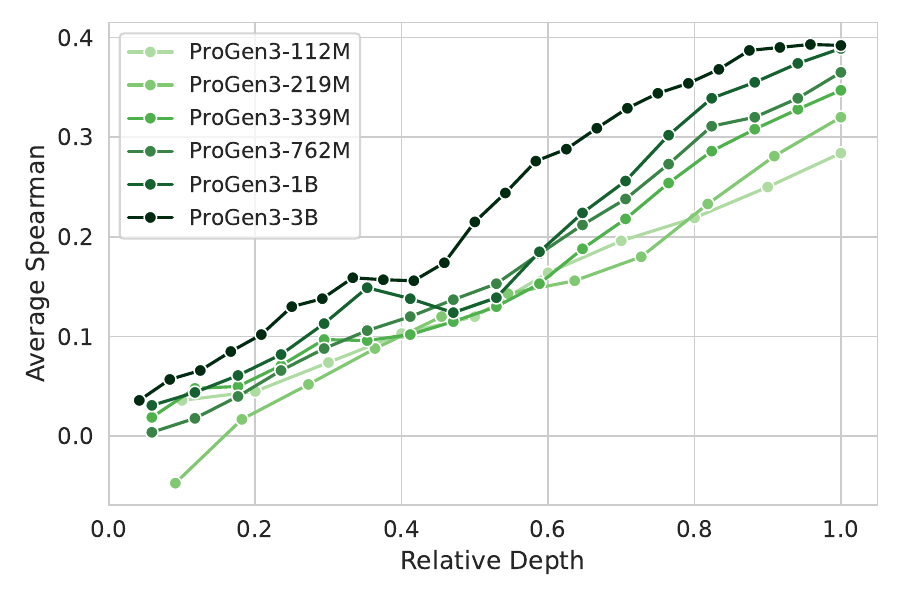}
    \caption{Average Spearman correlation for ProGen3 on ProteinGym, calculated at each layer. The relative depth is normalized to $[0,1]$.}
    \label{fig:layerwise_average_spearman_ProGen3}
\end{figure}

\cref{fig:layerwise_average_spearman_DPLM,fig:layerwise_average_spearman_DPLM2,fig:layerwise_average_spearman_E1,fig:layerwise_average_spearman_ESM3,fig:layerwise_average_spearman_ProGen2,fig:layerwise_average_spearman_ProGen3} report layer-wise ProteinGym performance of the models, measured as average Spearman correlation as a function of relative depth (normalized to $[0,1]$). These curves characterize where useful information for zero-shot fitness prediction is most strongly expressed in the representation stack: rising performance with depth suggests that later layers refine task-relevant signals, whereas early plateaus indicate that additional depth provides limited marginal improvement to the downstream performance.

\begin{figure}[h]
    \centering
    \includegraphics[width=0.98\textwidth]{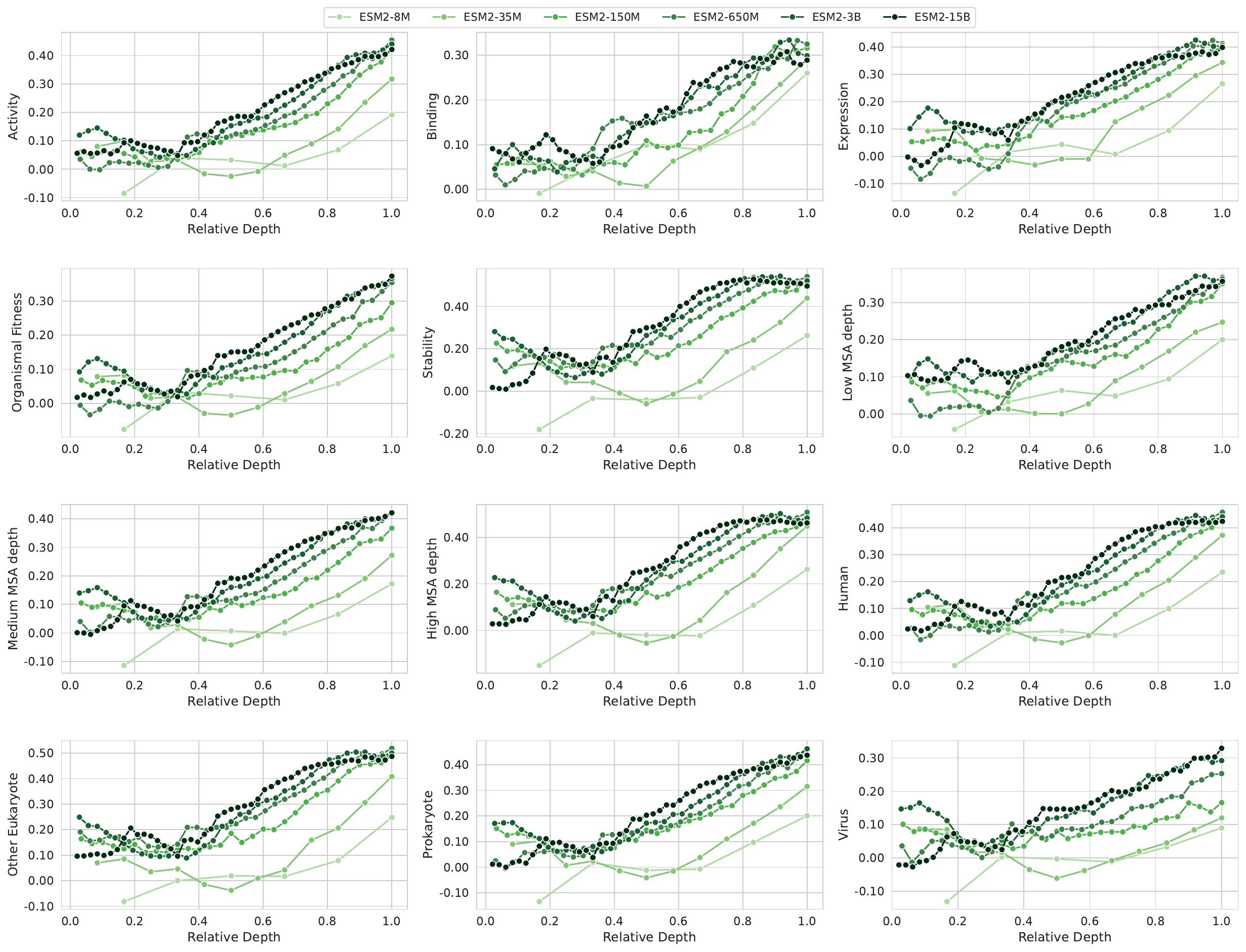}
    \caption{Average Spearman correlation for ESM2 on ProteinGym, computed at each layer and shown separately by phenotype. Relative depth is normalized to $[0,1]$.}
    \label{fig:layerwise_full_ESM2}
\end{figure}

\begin{figure}[h]
    \centering
    \includegraphics[width=0.98\textwidth]{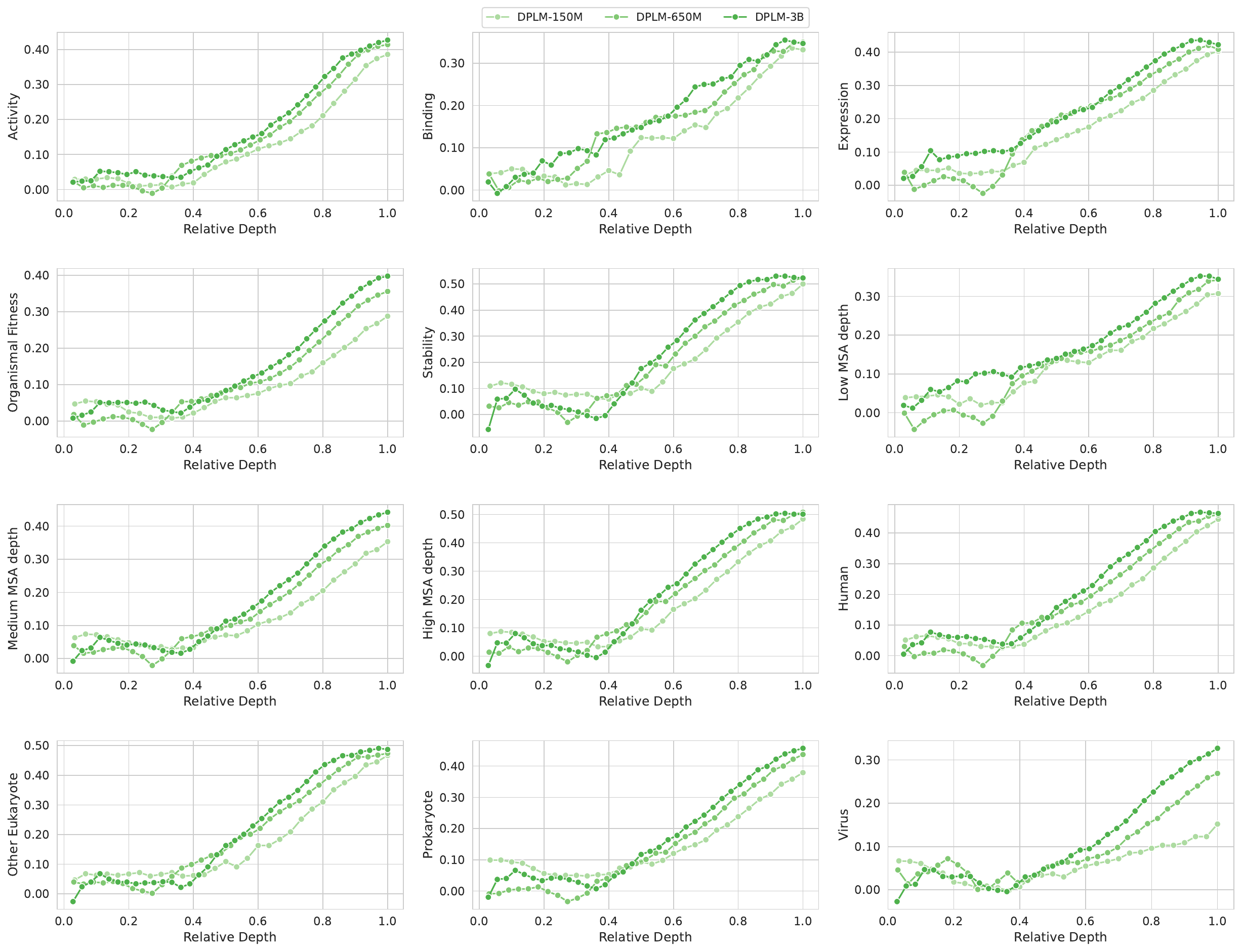}
    \caption{Average Spearman correlation for DPLM on ProteinGym, computed at each layer and shown separately by phenotype. Relative depth is normalized to $[0,1]$.}
    \label{fig:layerwise_full_DPLM}
\end{figure}

\begin{figure}[h]
    \centering
    \includegraphics[width=0.98\textwidth]{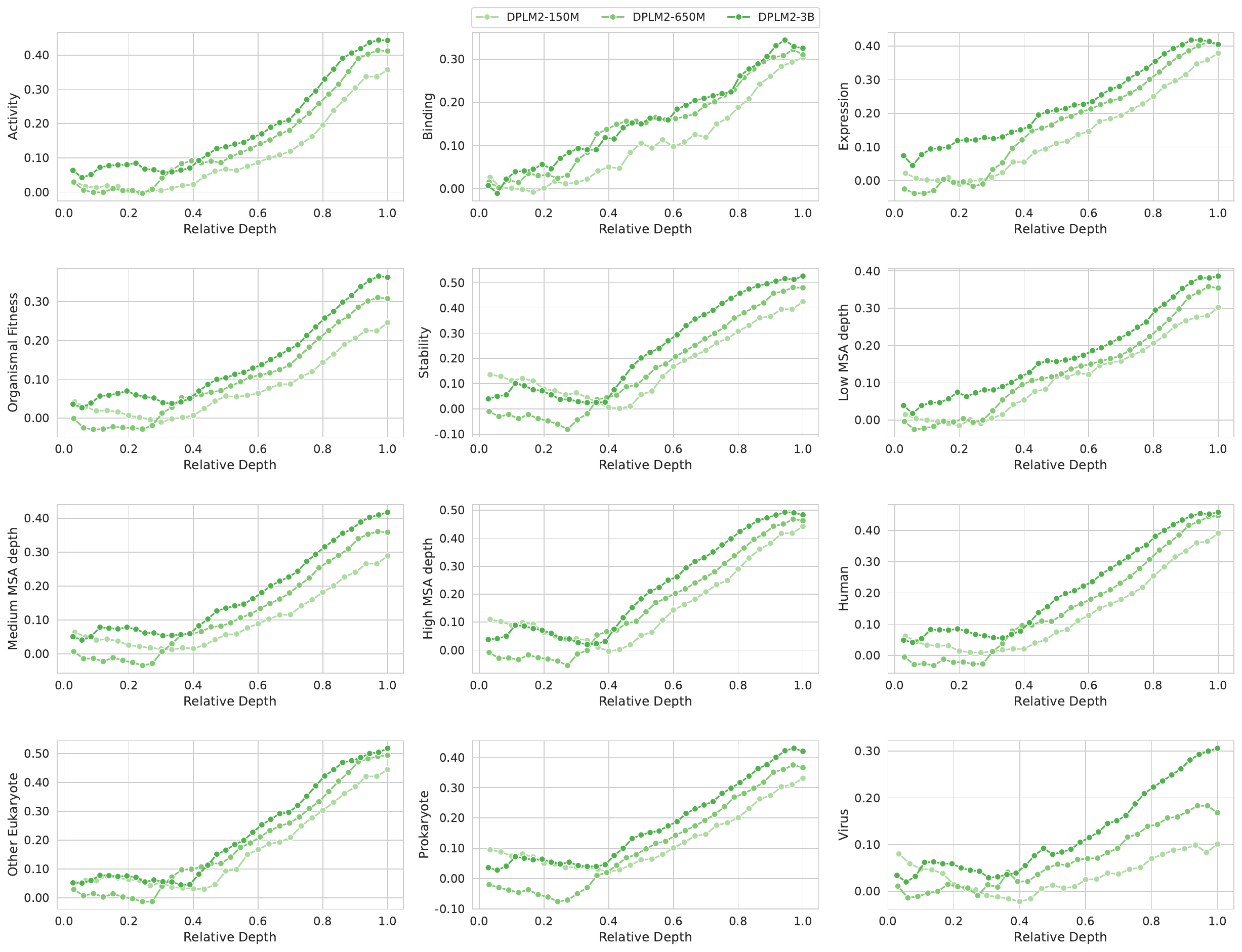}
    \caption{Average Spearman correlation for DPLM2 on ProteinGym, computed at each layer and shown separately by phenotype. Relative depth is normalized to $[0,1]$.}
    \label{fig:layerwise_full_DPLM2}
\end{figure}

\begin{figure}[h]
    \centering
    \includegraphics[width=0.98\textwidth]{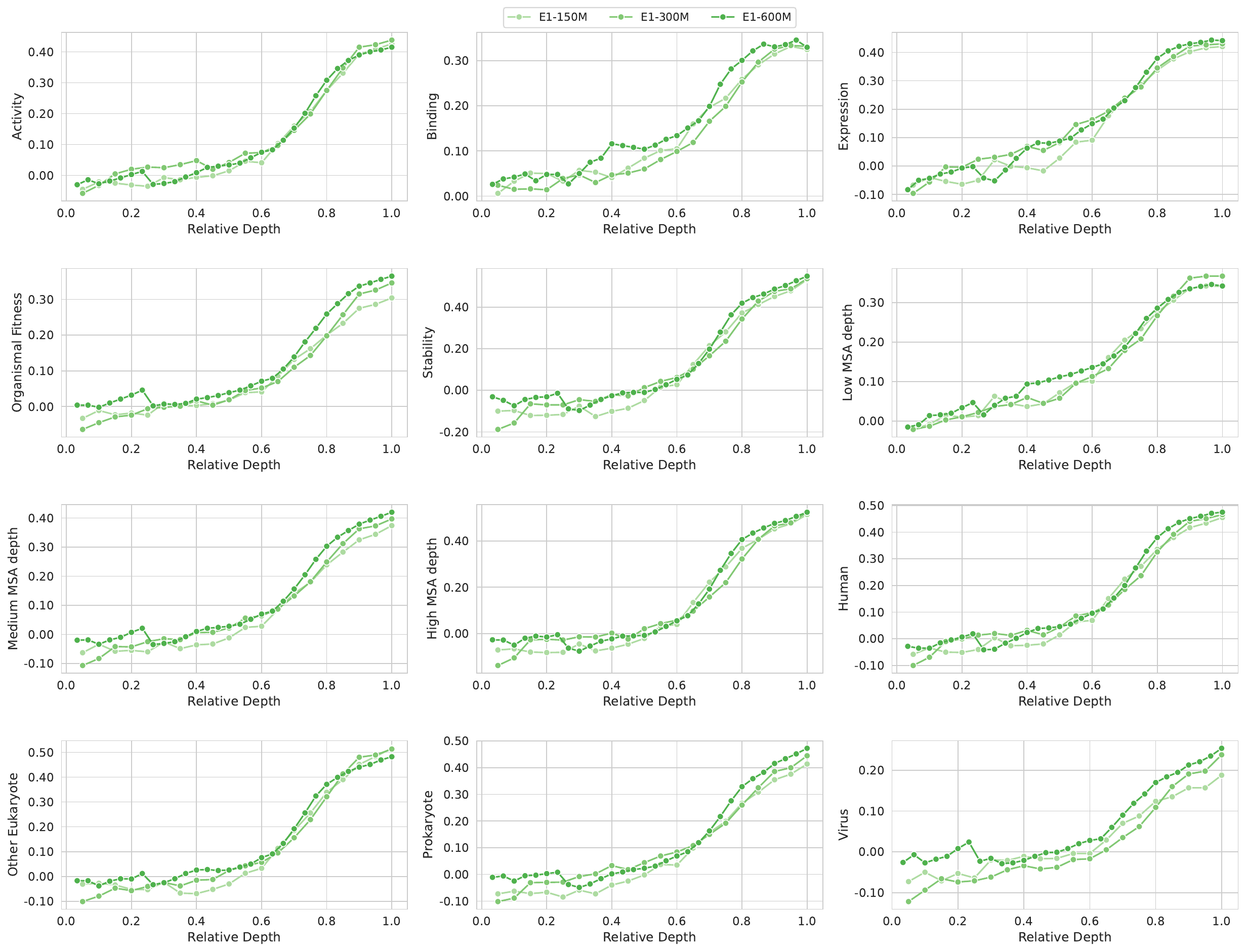}
    \caption{Average Spearman correlation for Profluent-E1 on ProteinGym, computed at each layer and shown separately by phenotype. Relative depth is normalized to $[0,1]$.}
    \label{fig:layerwise_full_E1}
\end{figure}

\begin{figure}[h]
    \centering
    \includegraphics[width=0.98\textwidth]{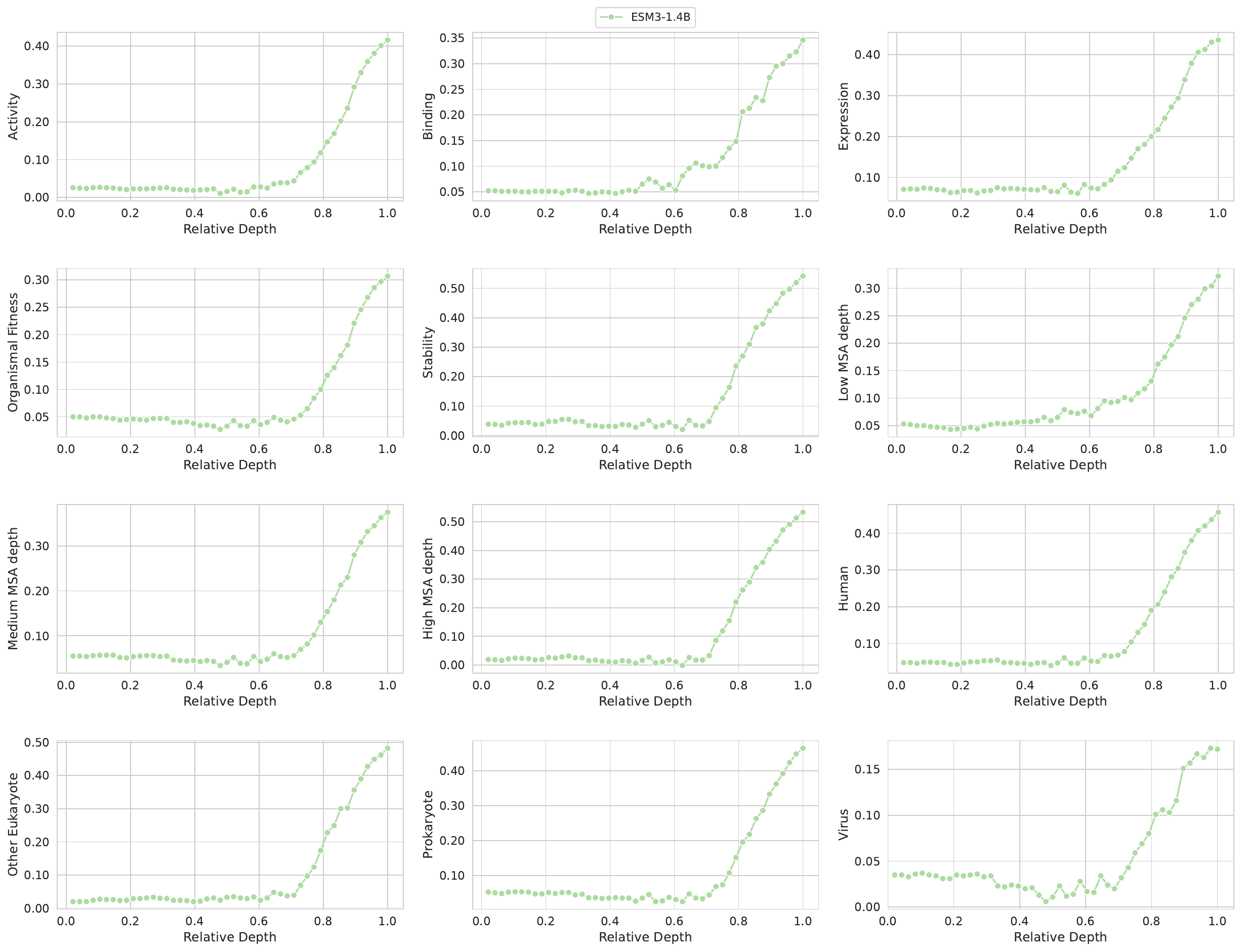}
    \caption{Average Spearman correlation for ESM3 on ProteinGym, computed at each layer and shown separately by phenotype. Relative depth is normalized to $[0,1]$.}
    \label{fig:layerwise_full_ESM3}
\end{figure}

\begin{figure}[h]
    \centering
    \includegraphics[width=0.98\textwidth]{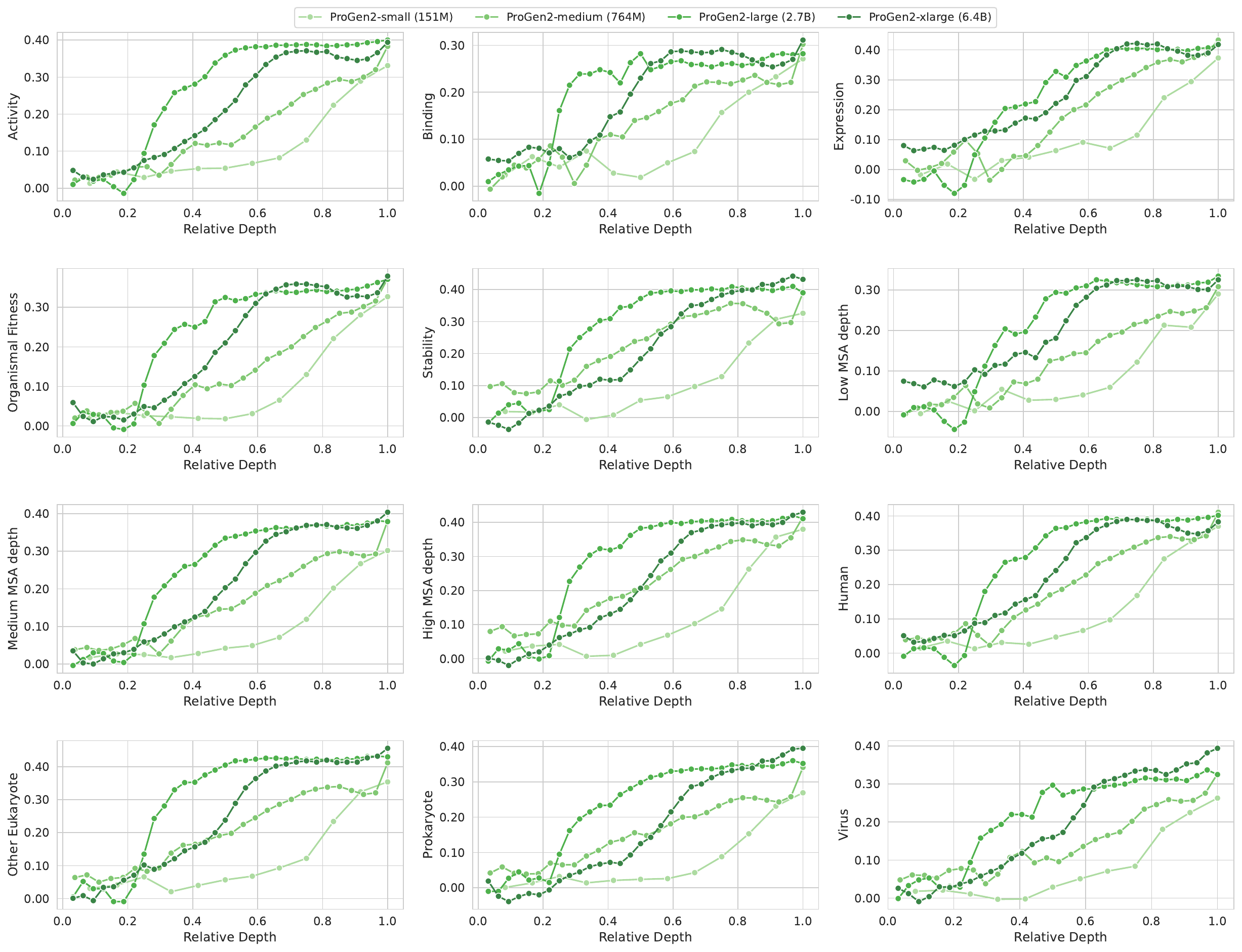}
    \caption{Average Spearman correlation for ProGen2 on ProteinGym, computed at each layer and shown separately by phenotype. Relative depth is normalized to $[0,1]$.}
    \label{fig:layerwise_full_ProGen2}
\end{figure}

\begin{figure}[h]
    \centering
    \includegraphics[width=0.98\textwidth]{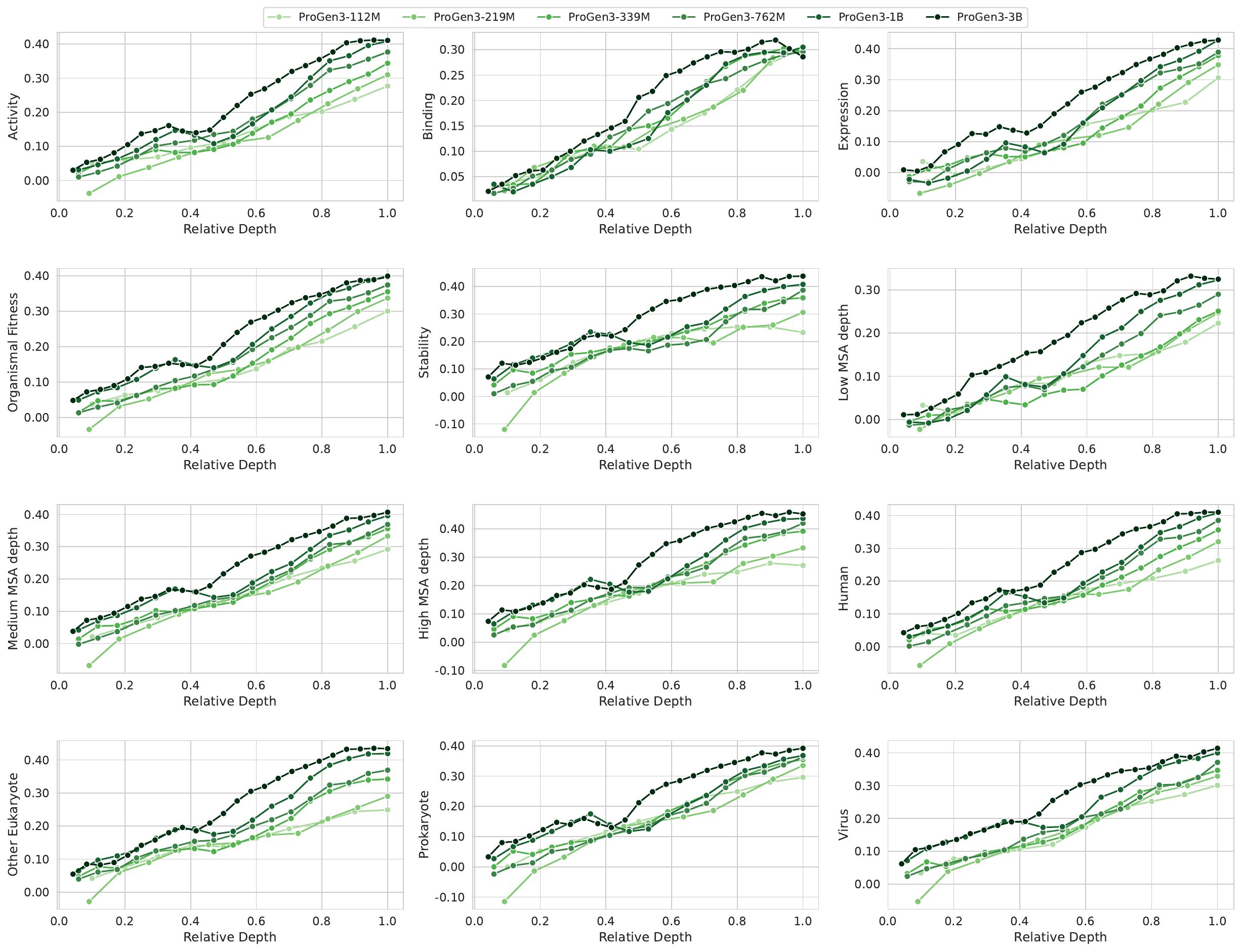}
    \caption{Average Spearman correlation for ProGen3 on ProteinGym, computed at each layer and shown separately by phenotype. Relative depth is normalized to $[0,1]$.}
    \label{fig:layerwise_full_ProGen3}
\end{figure}

To assess whether these trends are consistent across task types, we additionally report the same analysis broken down by phenotype in \cref{fig:layerwise_full_ESM2,fig:layerwise_full_ESM3,fig:layerwise_full_DPLM,fig:layerwise_full_DPLM2,fig:layerwise_full_E1,fig:layerwise_full_ProGen2,fig:layerwise_full_ProGen3}. The phenotype-level plots highlight which categories follow the overall average trend and which deviate by exhibiting earlier peaks or stronger reliance on deeper layers, providing a more fine-grained view of how depth affects downstream performance across different biological readouts.

Layer-wise downstream performance generally improves with depth but often exhibits diminishing returns in later layers for larger models, suggesting that much of the predictive signal is captured by intermediate depth and later layers provide incremental refinement. ESM3 shows a different depth allocation, with gains concentrated more strongly in later layers, consistent with its distinct intrinsic depth-wise patterns. When breaking results down by phenotype, most phenotype-specific curves closely follow the overall average trend. For a few phenotypes such as \textit{Stability}, the saturation is even clearer, but overall, the phenotype-level results support the same conclusion as the average.

\subsection{How does depth usage vary across modalities?}
\label{app:multimodal}

As stated in \cref{subsec:multimodal}, we test whether depth-inefficiency trends differ across modalities by repeating the same analyses from \cref{subsec:skiplayer_effects,subsec:lens} on multimodal PLMs under two additional settings: (i) \emph{structure-only}, where only structural tokens are provided as input, and (ii) a \emph{multimodal} setting, where both sequence and structure are provided. For DPLM2, sequence and structure share the same prediction head, so we use the same head in both settings. For ESM3, sequence and structure use separate prediction heads; accordingly, we use the structure head for the structure-only setting and the sequence head for the multimodal setting.

\begin{figure}[h]
    \centering
    \includegraphics[width=0.9\textwidth]{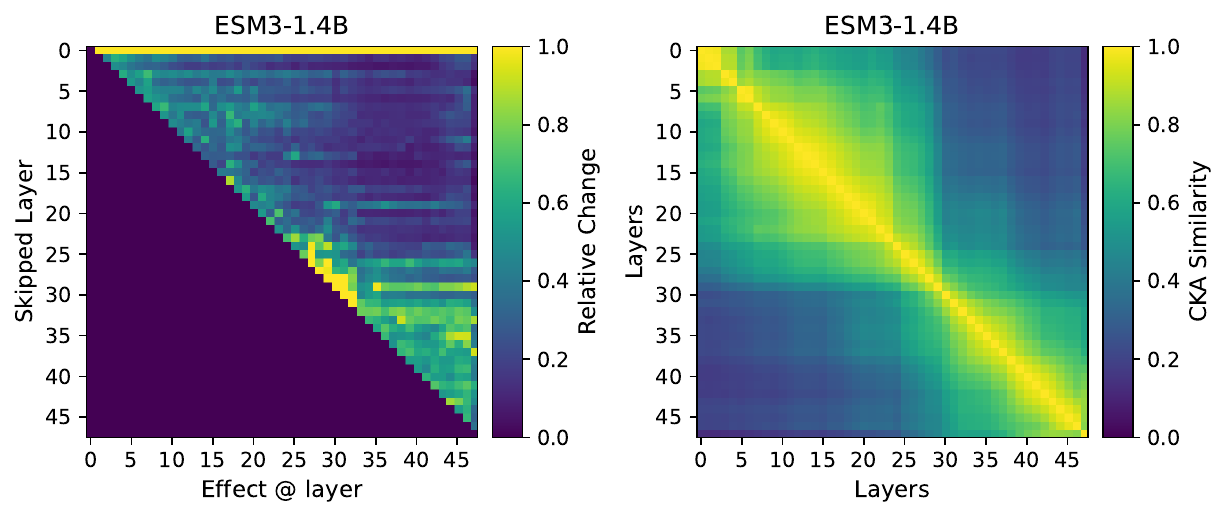}
    \caption{Maximum propagated effect of skipping each layer on future-token computations of the multimodal (sequence-structure) stream for ESM3.}
    \label{fig:skiplayer_layers_multimodal_ESM3}
\end{figure}

\begin{figure}[h]
    \centering
    \includegraphics[width=0.9\textwidth]{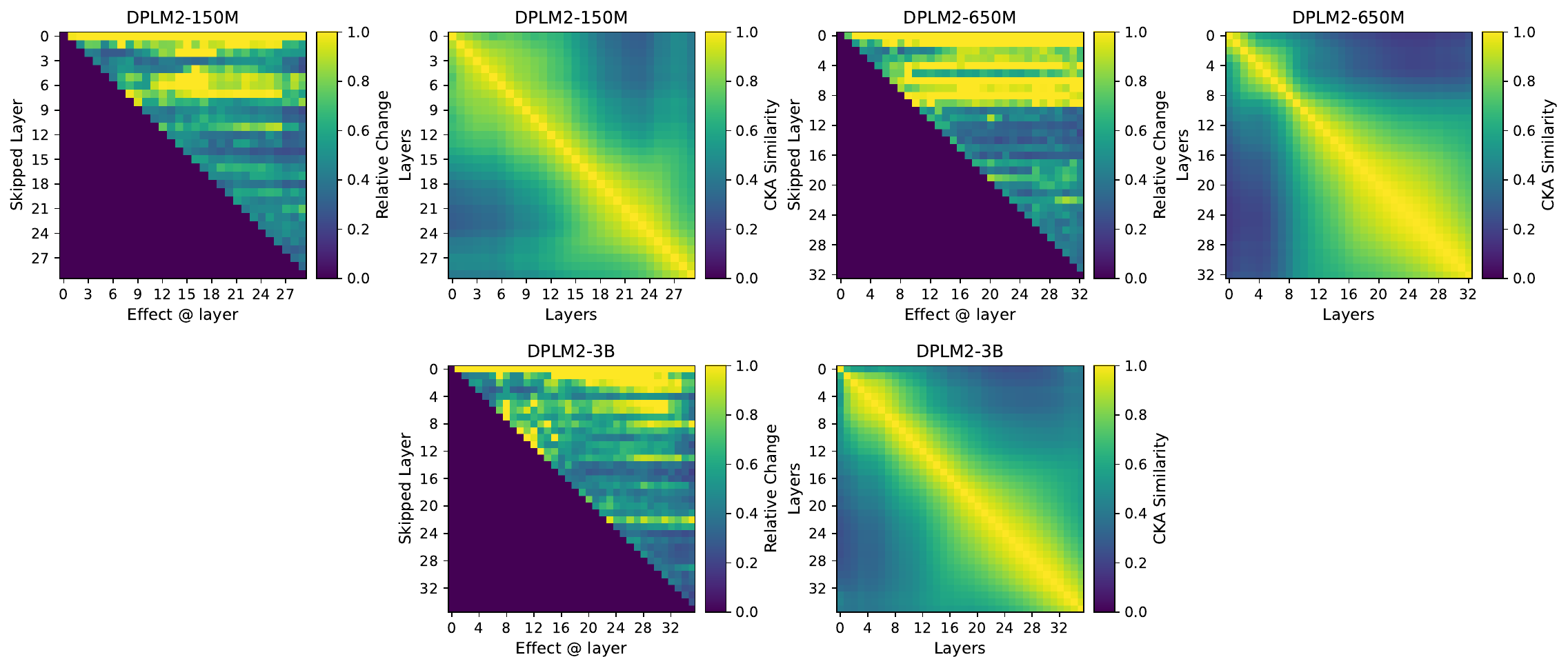}
    \caption{Maximum propagated effect of skipping each layer on future-token computations of the structure stream for DPLM2.}
    \label{fig:skiplayer_layers_structure_DPLM2}
\end{figure}

\begin{figure}[h]
    \centering
    \includegraphics[width=0.9\textwidth]{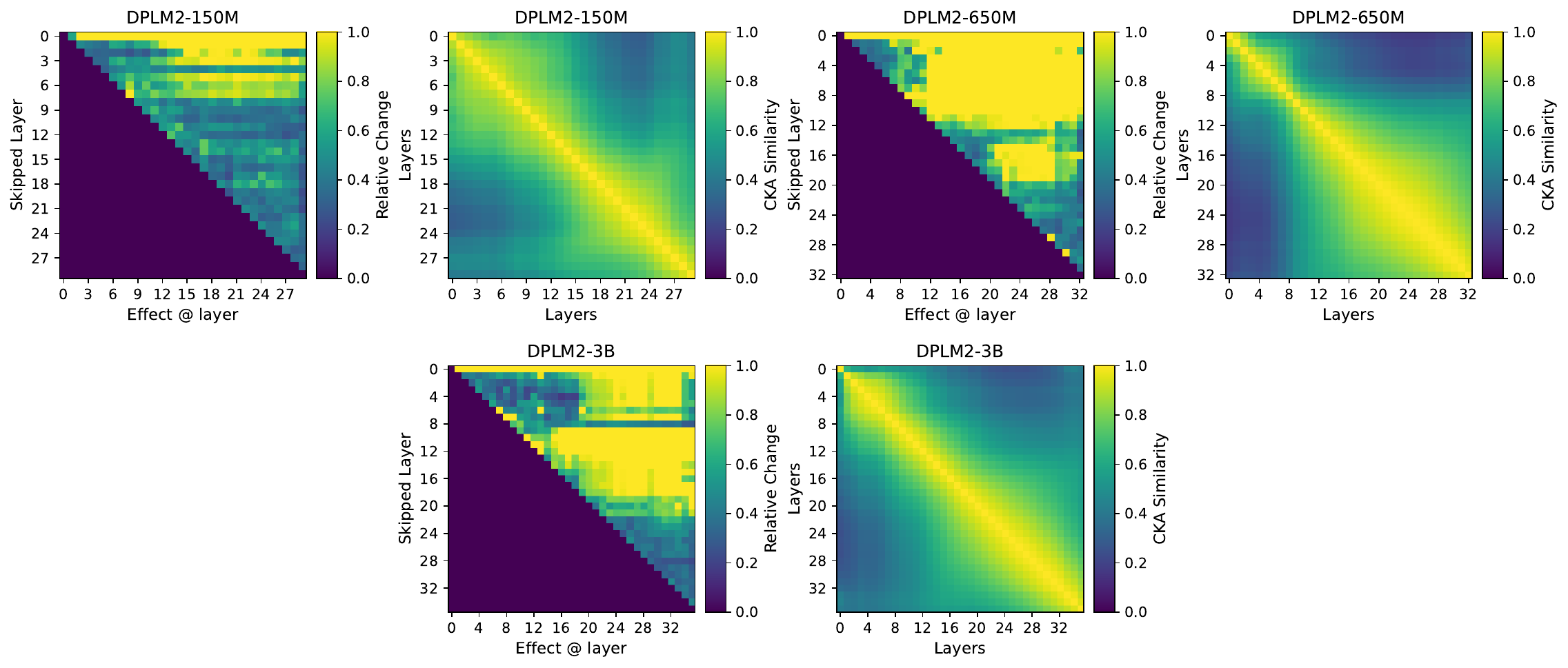}
    \caption{Maximum propagated effect of skipping each layer on future-token computations of the multimodal (sequence-structure) stream for DPLM2.}
    \label{fig:skiplayer_layers_multimodal_DPLM2}
\end{figure}

\begin{figure}[h]
    \centering
    \includegraphics[width=0.9\linewidth]{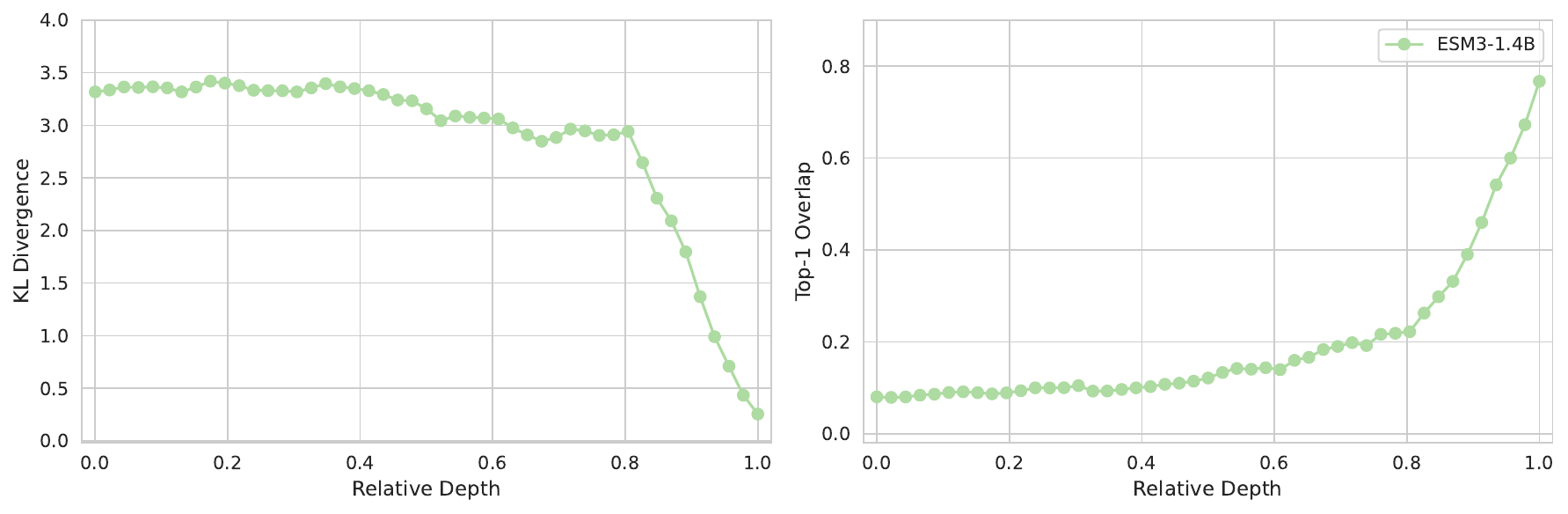}
    \caption{LogitLens analysis for ESM3 across depth on the multimodal (sequence-structure) stream: KL divergence between the layer-wise output distribution and the final output distribution (left), and top-1 overlap between the layer-wise prediction and the full-model prediction (right).}
    \label{fig:logitlens_multimodal_esm3}
\end{figure}

\begin{figure}[h]
    \centering
    \includegraphics[width=0.9\linewidth]{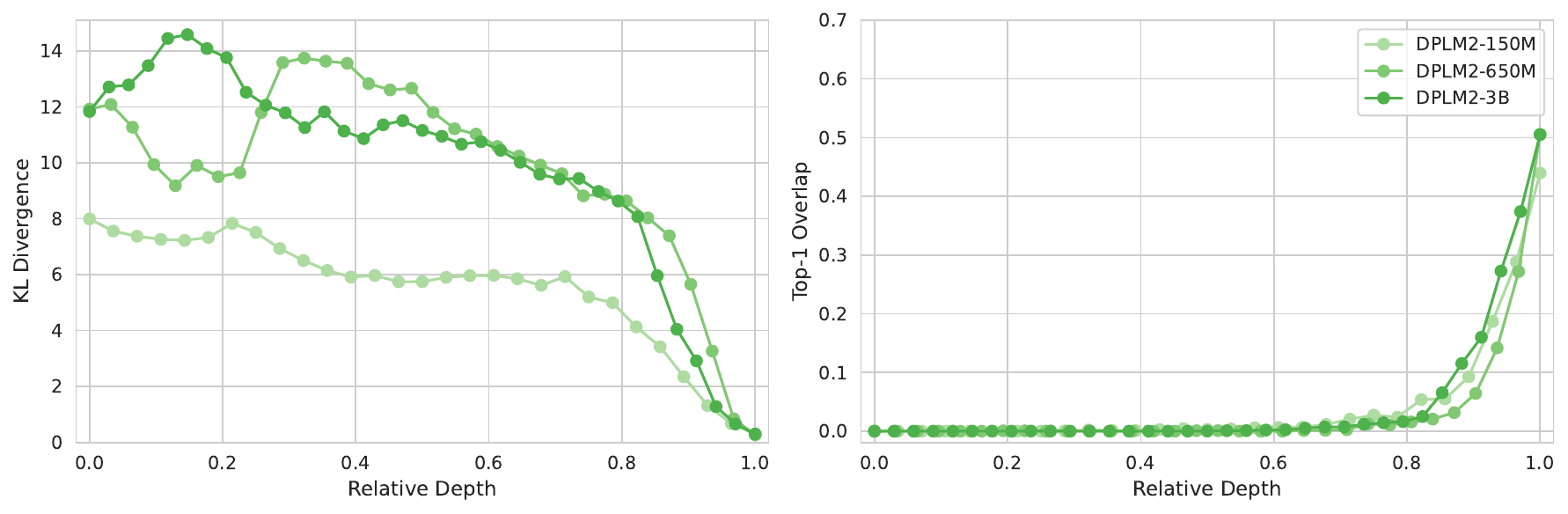}
    \caption{LogitLens analysis for DPLM2 across depth on the structure stream: KL divergence between the layer-wise output distribution and the final output distribution (left), and top-1 overlap between the layer-wise prediction and the full-model prediction (right).}
    \label{fig:logitlens_structure_dplm2}
\end{figure}

\begin{figure}[h]
    \centering
    \includegraphics[width=0.9\linewidth]{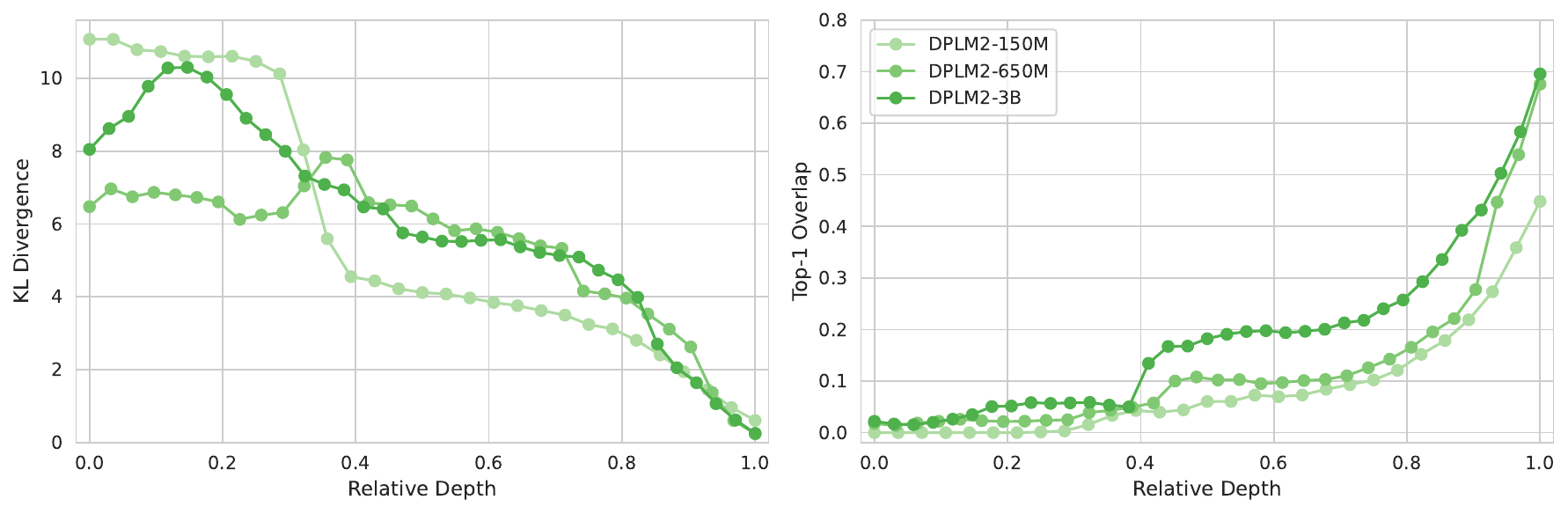}
    \caption{LogitLens analysis for DPLM2 across depth on the multimodal (sequence-structure) stream: KL divergence between the layer-wise output distribution and the final output distribution (left), and top-1 overlap between the layer-wise prediction and the full-model prediction (right).}
    \label{fig:logitlens_multimodal_dplm2}
\end{figure}

\cref{fig:skiplayer_layers_structure_DPLM2,fig:logitlens_structure_dplm2} show the structure-only analysis results for DPLM2. The multimodal analysis results for ESM3 are shown in \cref{fig:skiplayer_layers_multimodal_ESM3,fig:logitlens_multimodal_esm3} and those for DPLM2 are shown in \cref{fig:skiplayer_layers_multimodal_DPLM2,fig:logitlens_multimodal_dplm2}. Across both models, we observe the same high-level trends as in their respective sequence pathway results. For DPLM2, both the structure-only setting and the multimodal setting show a sharp separation between early/mid layers with strong propagated effects and late layers with much weaker effects. Complementarily, LogitLens results show KL divergence decreasing and top-1 overlap increasing toward later layers. For ESM3, in both settings, we again observe a low-effect region in the early layers and then again a weaker low-effect region in the later layers. Complementarily, KL divergence remains relatively high through early-to-mid depth and then drops sharply near the end, while top-1 overlap rises most strongly in the final layers.

\begin{figure}[h]
    \centering
    \includegraphics[width=0.9\linewidth]{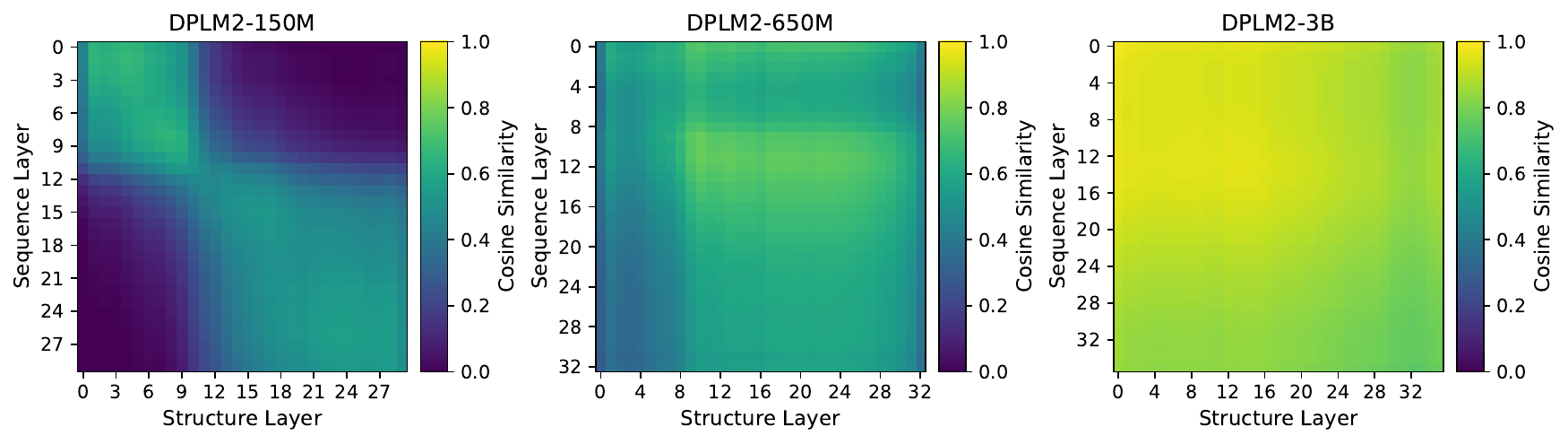}
    \caption{Cross-modal layer--layer cosine similarity between sequence-only and structure-only representations for DPLM2, averaged across proteins. Each cell compares a sequence-layer representation to a structure-layer representation for the same proteins.}
    \label{fig:crossmodal_similarity_dplm2}
\end{figure}

To complement the analysis in \cref{subsec:esm3_modality_alignment}, we report cross-modal layer-layer cosine similarity between sequence-only and structure-only representations for ESM3 and DPLM2. For each protein, we compare the representation from each sequence layer with the representation from each structure layer, and average the similarities across proteins. \cref{fig:crossmodal_similarity_esm3} shows that ESM3 has a clear two-phase pattern, with high cross-modal similarity in the first $\sim$60\% of layers followed by a more localized similarity structure in later layers. In contrast, \cref{fig:crossmodal_similarity_dplm2} shows a more uniform cross-modal similarity pattern for DPLM2, without the same sharp early alignment-to-divergence transition. These results support the interpretation that ESM3 devotes early depth to sequence-structure alignment more strongly than DPLM2.

\end{document}